\documentclass{article} 
\usepackage{iclr2026_conference,times}
\usepackage{graphicx}
\usepackage{subcaption} 

\usepackage{tikz}
\usepackage{pgfplots}
\pgfplotsset{compat=newest}

\usepackage{svg}
\svgpath{{svg/}} 
\svgsetup{
  inkscapelatex=true, 
  inkscapeversion=auto  
}

\usepackage{amsmath,amsfonts,bm}

\def\eqref#1{equation~\ref{#1}}

\def\1{\bm{1}}

\DeclareMathAlphabet{\mathsfit}{\encodingdefault}{\sfdefault}{m}{sl}
\SetMathAlphabet{\mathsfit}{bold}{\encodingdefault}{\sfdefault}{bx}{n}

\usepackage{hyperref}
\usepackage{url}
\usepackage{xcolor}

\usepackage{listings} 
\usepackage{booktabs} 
\usepackage{siunitx}
\usepackage{amsmath}

\usepackage{microtype}
\usepackage{lmodern}
\title{TRELLISWorld: Training-Free World Generation from Object Generators}

\author{
Hanke Chen\textsuperscript{1}\ \ \ \ \ \ 
Yuan Liu\textsuperscript{2} \ \ \ \ \ \
Minchen Li\textsuperscript{1}\\
\textsuperscript{1}Carnegie Mellon University\ \ \ \ \ \ 
\textsuperscript{2}The Hong Kong University of Science and Technology \\
\texttt{\{hankec, minchenl\}@andrew.cmu.edu}\ \ \ \ \ \ 
\texttt{yuanly@connect.ust.hk} 
}

\iclrfinalcopy 
\begin{document}

\maketitle

\begin{figure}[h]
  \centering
  \includegraphics[width=\textwidth]{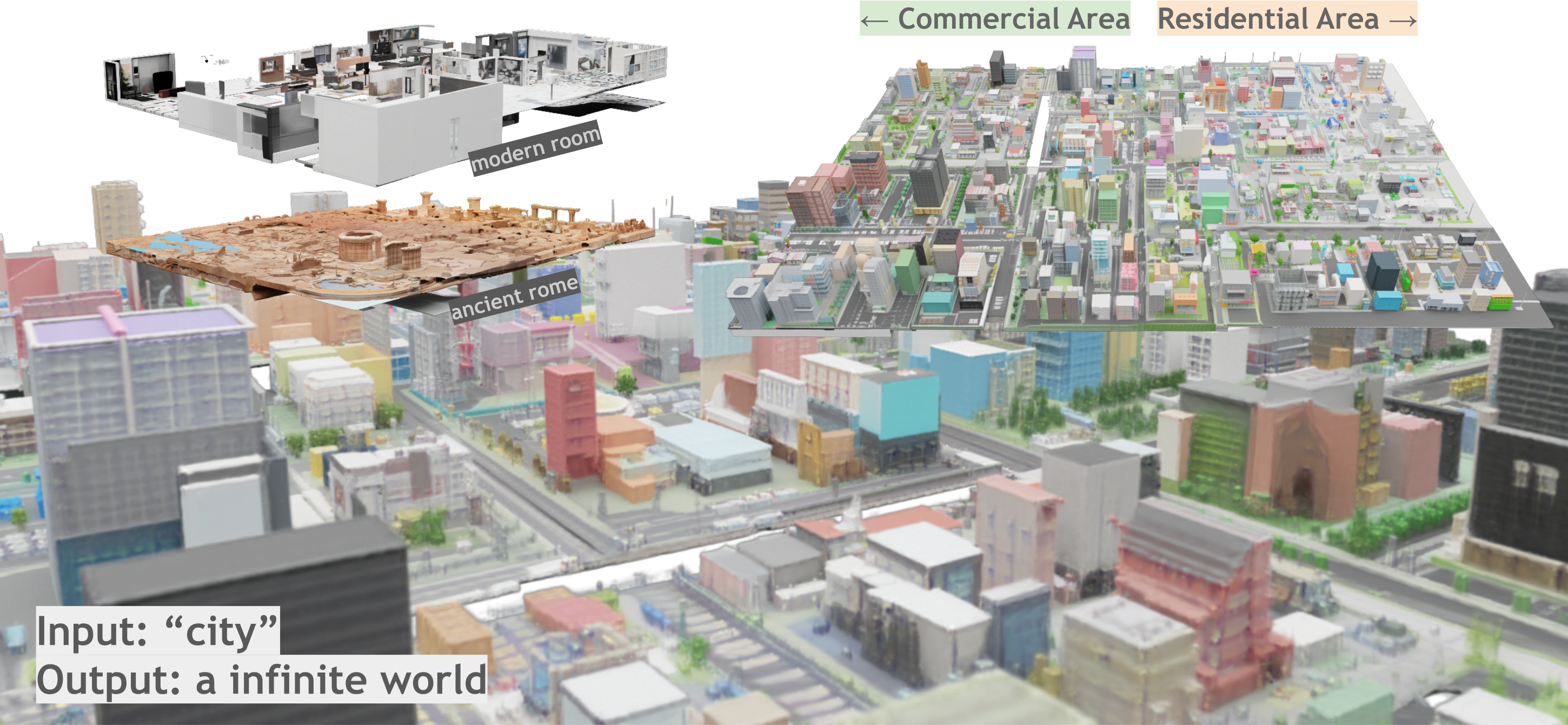}
  \caption{Scenes generated by our framework, \textbf{TRELLISWorld}, using only natural language input. Users may provide fine-grained prompts for specific regions, enabling semantically consistent gradual transitions: e.g., from a dense commercial district with greens, into low-density residential zones.}
  \label{fig:teaser}
\end{figure}

\begin{abstract}
Text-driven 3D scene generation holds promise for a wide range of applications, from virtual prototyping to AR/VR and simulation. However, existing methods are often constrained to single-object generation, require domain-specific training, or lack support for full 360-degree viewability. In this work, we present a training-free approach to 3D scene synthesis by repurposing general-purpose text-to-3D object diffusion models as modular tile generators. We reformulate scene generation as a multi-tile denoising problem, where overlapping 3D regions are independently generated and seamlessly blended via weighted averaging. This enables scalable synthesis of large, coherent scenes while preserving local semantic control. Our method eliminates the need for scene-level datasets or retraining, relies on minimal heuristics, and inherits the generalization capabilities of object-level priors. We demonstrate that our approach supports diverse scene layouts, efficient generation, and flexible editing, establishing a simple yet powerful foundation for general-purpose, language-driven 3D scene construction. We will release the full implementation upon publication.
\end{abstract}

\section{Introduction}

Generating 3D worlds from text represents a longstanding goal at the intersection of computer graphics, machine learning, and human-computer interaction. Enabling users to describe a virtual world in natural language and synthesize an editable 3D environment would transform multiple domains. For example, in creative content design, this could accelerate ideation workflows, allowing a game designer to rapidly prototype a level using coarse spatial prompts, or enabling a video creator to synthesize a virtual scene as a storytelling background.

Recent advances in 3D content generation, ranging from SDS-based methods distilling 2D priors into neural fields \citep{sds, vsd, sjc, dreamgaussian}, to multi-view diffusion for geometric consistency \citep{z123, z123pp, mvdream, syncdreamer}, and emerging 3D-native models like DiTs \citep{dit, lrm, hunyuan3d, clay}, have significantly improved 3D object synthesis. However, these methods are predominantly limited to single-object generation, with limited progress toward generating entire 3D scenes.

While directly generating 3D scenes is an active research direction, current scene-generation models tend to be either (1) domain-specific, trained on narrow datasets like indoor rooms or driving scenes; or (2) produce image-based representations such as panoramic or spherical projections, which are not designed for full 360-degree spatial interaction or view synthesis. This limitation arises from limited large-scale, diverse, general-purpose datasets for 3D scenes comparable to those available in 2D vision or object-centric 3D generation. Scenes contain long-range dependencies, heterogeneous object types, and complex spatial arrangements that are difficult to annotate and curate. Therefore, a method that can leverage object-centric generation to generate scenes is highly desired.

Recently, SynCity \citep{syncity} demonstrates promising results in adapting object generators for city-scale scene generation. However, its dependence on 2D inpainting \citep{inpaint} introduces a fundamental limitation. Errors in the image domain can propagate and destabilize the 3D reconstruction. This raises a natural question: can we instead generate scenes directly in 3D space, bypassing the need for 2D intermediate representations?

In this work, we introduce \textbf{TRELLISWorld}, a training-free approach to text-driven 3D scene generation by leveraging general-purpose text-to-3D object diffusion models for scene composition. Our key insight is to reformulate global 3D scene synthesis as a multi-tile denoising problem, wherein the scene is partitioned into spatially overlapping regions that are independently denoised and later blended using a weighted averaging scheme in one diffusion step. This formulation offers a practical and scalable alternative to end-to-end scene-level training, enabling high-quality scene generation at significantly reduced cost. We will release the full implementation upon publication. Our method offers several advantages:
\begin{itemize}
    \item \textbf{Training-free and editable}: Taking advantage of 3D environments' multi-scale signal structure, our approach requires no scene-level dataset or fine-tuning. It inherits editability and generalization capabilities from the underlying object-level generator, e.g., TRELLIS \citep{trellis}.
    \item \textbf{Simple and general}: Our method requires minimal task-specific heuristics, making it broadly applicable across diverse scene types.
    \item \textbf{Scalable and smooth}: Compared to prior methods, our tile-wise approach is computationally efficient, blends overlapping regions smoothly, and enables the generation of significantly larger and more coherent scenes.
\end{itemize}

\section{Related Work}

\subsection{Foundation of Reconstruction and Object Generation}
Benefiting from recent advances in 3D representations for reconstruction, such as NeRF \citep{nerf, plenoxel, ngp, triplane, tensorf, gsplat, nerfpp, mipnerf, mipnerf360, zipnerf}, Score Distillation Sampling (SDS)-based methods \citep{sds, vsd, sjc, dreamgaussian} distill the knowledge of 2D diffusion models into the creation of 3D objects using differentiable rendering. Specifically, NeRF++ \citep{nerfpp}, combined with ProlificDreamer \citep{vsd}, demonstrates the first possibility of generating 3D scenes by distilling 2D diffusion with an added density prior. Later, multi-view diffusion methods \citep{z123, z123pp, mvdream, syncdreamer} were proposed to address the Janus (multi-face) problem. However, as SDS depends on test-time optimization, object generation can take up to 40 minutes. To overcome this, methods like LRM have been proposed to generate 3D representations directly from images \citep{12345, lrm, cat3d, hunyuan3d} and text \citep{shape-e, 3dtopia, clay}. Although none of these works aim to generate large-scale scenes, their technologies lay the foundation for 3D scene generation.

\subsection{Scene Generation Based on 2D Generation}
While SDS-based methods above often focus on novel view synthesis or scene reconstruction, another line of work leverages the capabilities of 2D diffusion models without relying on test-time optimization. Early works extend a single image autoregressively into a video sequence, guided by camera trajectories \citep{infinite-nature, infinite-nature-zero, persist-nature, diffdreamer}. These methods often project depth-estimated results into point clouds and fill in the missing regions under different camera extrinsics. While initial work focused mainly on natural scenes, the introduction of generic 2D diffusion models expanded the domain \citep{scenescape, realmdreamer, luciddreamer-c}, and the integration of large language models (LLMs) enabled more diverse scene generation \citep{wonderjourney, wonderworld, hunyuanworld}. However, these works are limited to generating panoramic images or incomplete 3D representations that can only be viewed from a restricted set of camera extrinsics. A generic 3D generator capable of producing complete meshes remains necessary.

\subsection{Scene Generation Based on 3D Scene-Native Generation}
3D-native generation methods are promising alternatives to produce complete 3D representations, but they often rely on domain-specific training and are unable to generate generic scenes using natural language prompts \citep{semcity, diffuscene, scenegen, nuiscene}. For instance, works like InfiniCity \citep{infinicity, gaussiancity, citydreamer4d} depend heavily on RGB-D, semantic, or normal maps derived from satellite imagery. CityDreamer4D \citep{citydreamer4d} decomposes the generation task into multiple sub-tasks: layout, background, buildings, vehicles, and roads, each handled by a separate neural network. BlockFusion \citep{blockfusion} uses autoregressive inference, while MIDI \citep{midi} employs multi-instance attention to scale object-level generation to scenes, both trained on dedicated 3D datasets \citep{3dfront, 3dfuture}. In contrast to publicly accessible 3D object datasets like Objaverse \citep{objaverse, objaverse-xl}, which contain over 10 million internet-sourced objects, the largest 3D scene dataset, FurniScene, contains only 100k rooms with 89 object classes \citep{furniscene}. To the best of our knowledge, no openly available generic 3D scene dataset currently exists. Thus, a method that does not rely on curated 3D scene datasets but can still generate 3D representations across domains is highly desirable.

\subsection{Scene Generation Based on 3D Object Generation}
There are two categories of works that utilize 3D object generators to build scenes. The first category focuses on generating individual 3D objects and uses LLMs, visual-language models (VLMs), or image-based techniques to infer plausible object positions and orientations \citep{layoutgpt, diorama, phip-g}. GALA3D \citep{gala3d} employs SDS and additional physical losses to refine LLM-generated scene layouts from textual descriptions. CAST \citep{cast} constructs a relational graph, a constraint graph, and multiple masks from a single image to generate both object instances without occlusion and their spatial arrangement, using physical losses for consistency. However, these methods either depend on image inputs, limiting composition outside the view frustum, or rely on LLM reasoning, which often fails to produce accurate coordinates and complex inter-object relationships.

The second category relies on the inpainting capability of 3D diffusion models to compose and blend scenes from object-level generations. Most relevant to our work, SynCity \citep{syncity} generates object chunks autoregressively through a loop of 2D inpainting, 3D generation, and rendering. After generation, it fixes seams between chunks using 3D inpainting. To ensure consistency between 3D chunks and to avoid occlusions during 2D inpainting, multiple heuristics, such as cutting off parts of generated meshes for occlusion-free renderings, are applied. However, these heuristics reduce generalizability beyond urban scenes and increase failure cases.

\section{Method}

\begin{figure}[!t]
  \centering
  \includegraphics[width=\linewidth]{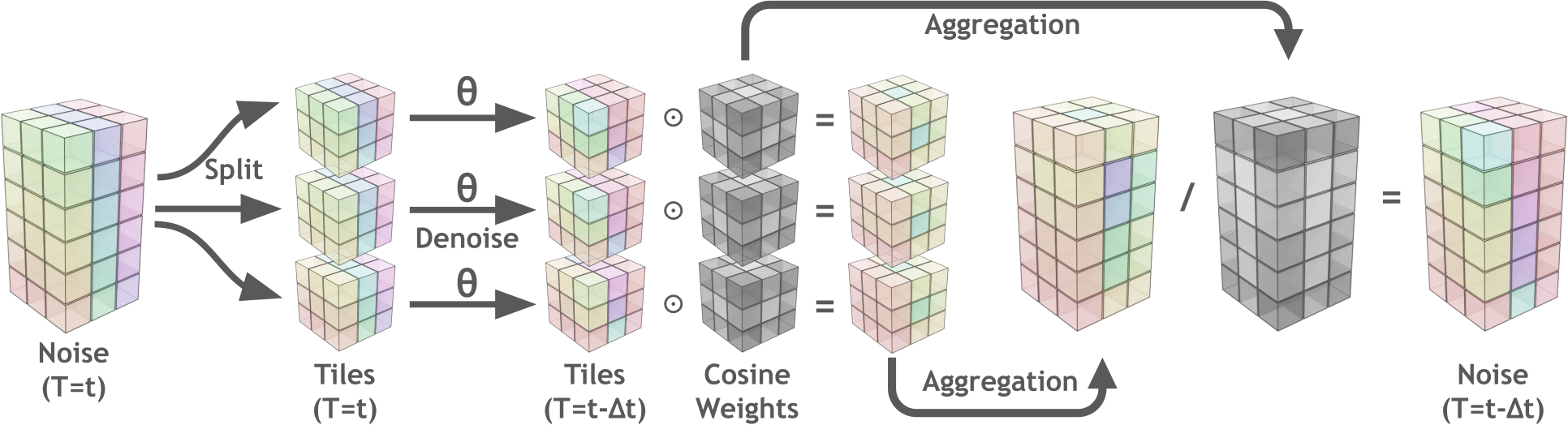}
  \caption{Illustration of our tiled diffusion process. We first split the scene noise into multiple tiles to denoise each tile in parallel. Then we take the weighted average described in \autoref{eq:weights} for each tile and aggregate the result to obtain the scene noise for previous timesteps. This process is detailed in \autoref{eq:tiled}.}
  \label{fig:tiled_diffusion}
\end{figure}

The core of TRELLISWorld is a tiled diffusion with cosine blending. We first formulate the problem in \autoref{subsec:problem_formulation}, then give a general method in \autoref{subsec:tiled_diffusion}. We then describe how we implement this method using TRELLIS in \autoref{subsec:implementation}.

\subsection{Problem Formulation}
\label{subsec:problem_formulation}

Given a text-conditioned 3D generative diffusion model $\theta$ (a velocity field \citep{flow-matching}) that is capable of generating a 3D structure of size $S^3$ from a text prompt $p$, our goal is to generate a large-scale 3D world of arbitrary size $(X \times Y \times Z) \gg S^3$ that is consistent with the prompt. To simplify our explanation, without loss of generality, we assume $\theta$ is a pixel-diffusion model, meaning that the forward and reverse diffusion processes operate on the actual values instead of on compressed latents produced by autoencoders. Therefore, each object sample can be represented with a tensor in $\mathbb{R}^{S^3}$. We further assume that $\theta$ is trained on a general object distribution. To generate worlds using an object-level 3D generator, we leverage the local generation capability of $\theta$ while ensuring global coherence through careful conditioning and blending techniques.

\subsection{Tiled Diffusion}
\label{subsec:tiled_diffusion}

In this section, we demonstrate a simple method to convert a general 3D object generator into a general 3D scene generator.

We first initialize the entire world $W$ of size $(X, Y, Z)$ with Gaussian noise $W \sim \mathcal{N}(\mathbf{0}, \mathbf{I})$. We then divide the world into overlapping cubic tiles $\{w_i\}$ of size $(S, S, S)$ with a stride of $(s, s, s)$, where $s < S$ to ensure overlap between adjacent tiles. For a fixed scene, different settings of $s$ give different numbers of tiles. Thus, we use the word "chunk" to denote the absolute size of the scene. For each diffusion step, we process each tile $\{w_i\}$ in parallel and then aggregate the weighted average result to update the world $W$. The weight for each tile is defined by a 3D cosine mask that emphasizes the center of the tile and tapers off towards the edges, ensuring smooth transitions between adjacent tiles. See \autoref{fig:tiled_diffusion} for visualization.

Formally, let $v^{(x, y, z)}$ be a voxel that has global position $(x, y, z)$ and let $f_{w_i}: \mathbb{N}^3 \to \mathbb{Z}^3$ be a function mapping global positions into local positions relative to tile $w_i$. Intuitively, if the resulting position $f_{w_i}(\cdot)$ lies in the set $\{0, ..., S - 1\}^3$, then the tile $w_i$ covers the voxel at this input global position.

We define the weighting on local position $(x', y', z')$ to be:

\begin{equation}
    \beta(x', y', z') = 
    \begin{cases}
        \displaystyle\prod_{d \in \{x', y', z'\}} \cos\left( \pi \left( \frac{d+1}{S+1} - \frac{1}{2} \right) \right) & \text{if } (x', y', z') \in \{0, \ldots, S - 1\}^3 \\
        0 & \text{otherwise}
    \end{cases}
    \label{eq:weights}
\end{equation}

Then, the update rule for each voxel $v^{(x, y, z)}$ in the world $W$ at diffusion step $t$ is given by the formula:

\begin{equation}
    v^{(x, y, z)}_{t - \Delta t} = \frac{\sum_{\{w_i^{(t)}\}} \beta(f_{w_i^{(t)}}(x, y, z)) \cdot \left[w_i^{(t)}-\Delta t \cdot \theta(w_i^{(t)}, t) + \mathcal{O}(\Delta t^2)\right]_{f_{w_i^{(t)}}(x, y, z)}}{\sum_{\{w_i^{(t)}\}} \beta(f_{w_i^{(t)}}(x, y, z))}
    \label{eq:tiled}
\end{equation}

where $\mathcal{O}(\Delta t^2)$ denotes the Big-O error resulting from Euler discretization, and the subscript $\left[\cdot\right]_{f_{w_i^{(t)}}(x, y, z)}$ denotes that $\theta$ operates in the local tile space.

\subsection{Implementation}
\label{subsec:implementation}

We build our method based on TRELLIS-text \citep{trellis}, which is a text-conditioned diffusion transformer for 3D object generation. To perform object-level inference with TRELLIS, the process begins by denoising a $16^3$ noised latent using the \textit{TRELLIS structure diffusion transformer} $\theta_1$. The resulting latent is decoded by \textit{TRELLIS sparse structure decoder} to a $64^3$ occupancy grid (SS). This dense $64^3$ tensor is converted into a noisy sparse tensor by retaining only regions where the occupancy value exceeds zero. The noise $\sim \mathcal{N}(\mathbf{0}, \mathbf{I})$ is applied onto those regions. The sparse tensor is subsequently denoised using the \textit{TRELLIS structure latent diffusion transformer} $\theta_2$, resulting in a structured latent representation (SLAT). Finally, the SLAT is decoded using the \textit{TRELLIS structure latent Gaussian decoder} to produce a Gaussian Splatting representation \citep{gsplat}.

\paragraph{Tiled Diffusion} As described above, TRELLIS is a multi-stage latent diffusion model where $\theta_1$ operates on dense tensors and $\theta_2$ operates on sparse tensors. We perform tiled diffusion with minimal modification on both stages: the input tiles to $\theta_1, \theta_2$ are in encoded latent space and the corresponding masks are down-sampled by $4 \times$. The blending and aggregation steps described in \autoref{subsec:tiled_diffusion} are performed in the latent space. 

\paragraph{Tiled Decoder} The resulting latent is decoded back to the voxel space and/or Gaussian Splatting space after the diffusion process is complete. Importantly, such decoding should also be done in a tiled manner. Since \textit{TRELLIS structure latent Gaussian decoder} is not a probabilistic model, we set the stride $s = S$ to disable blending.

\section{Experiments}

We ablate our method in \autoref{subsec:ablation} and compare blend quality, perceptual alignments, and computational cost with SynCity in \autoref{subsec:qualitative} and \autoref{subsec:quantitative}. Unless otherwise noted, all experiments are conducted using classification-free guidance \citep{cfg} $cfg=7.5$, stride $s = \frac{S}{2}$ and diffusion step size $25$ on the Euler sampler.

\subsection{Ablation}
\label{subsec:ablation}

\begin{figure}[t]
    \centering
    \newcommand{\lblw}{0.005\textwidth}   
    \newcommand{\imgw}{0.25\textwidth}   
    \begin{tabular}{%
    >{\centering\arraybackslash}m{\lblw}%
    >{\centering\arraybackslash}m{\imgw}%
    >{\centering\arraybackslash}m{\imgw}%
    >{\centering\arraybackslash}m{\imgw}%
    }
    \rotatebox{90}{\scriptsize(a) Autoregressive} &
    \includegraphics[width=\imgw]{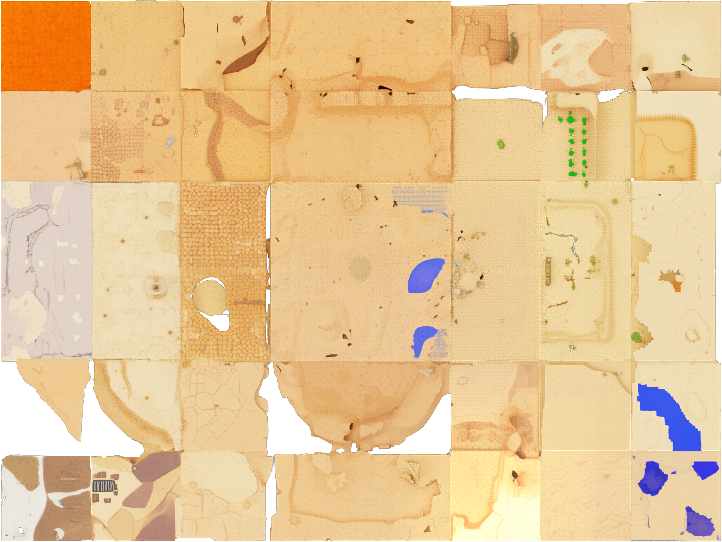} &
    \includegraphics[width=\imgw]{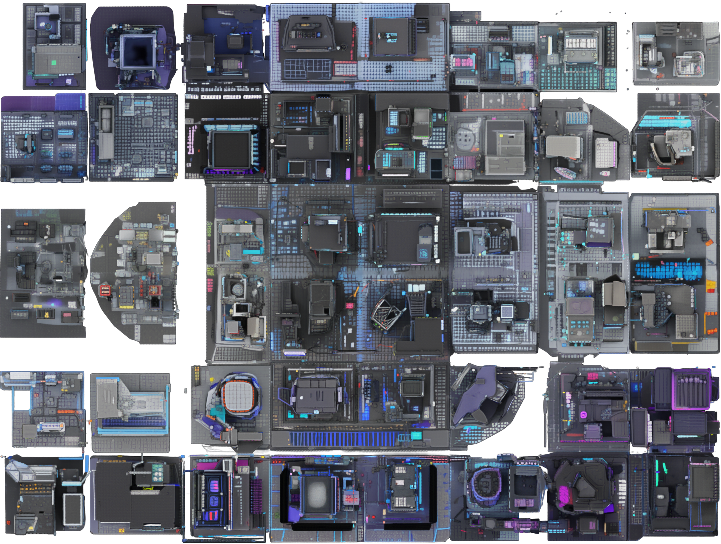} &
    \includegraphics[width=\imgw]{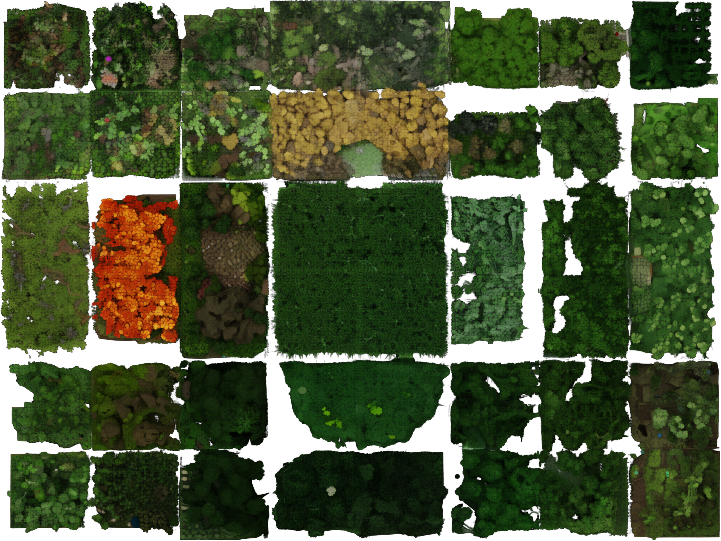}
    \\[4pt]
    \rotatebox{90}{\scriptsize(b) TRELLISWorld} &
    \includegraphics[width=\imgw]{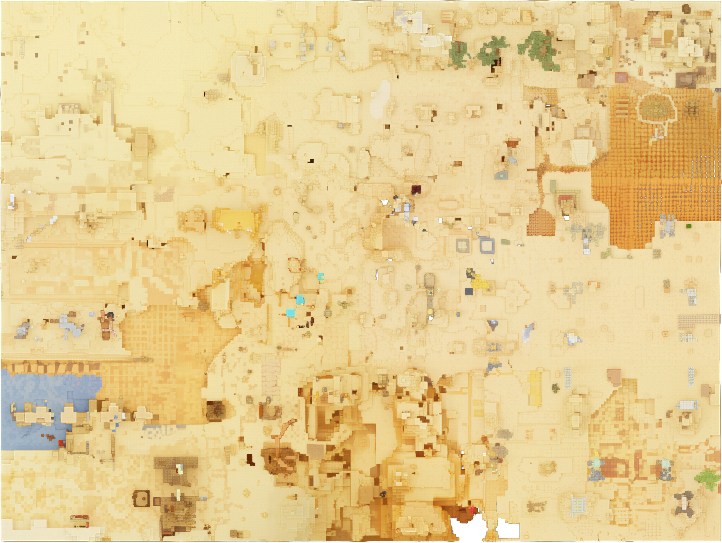} &
    \includegraphics[width=\imgw]{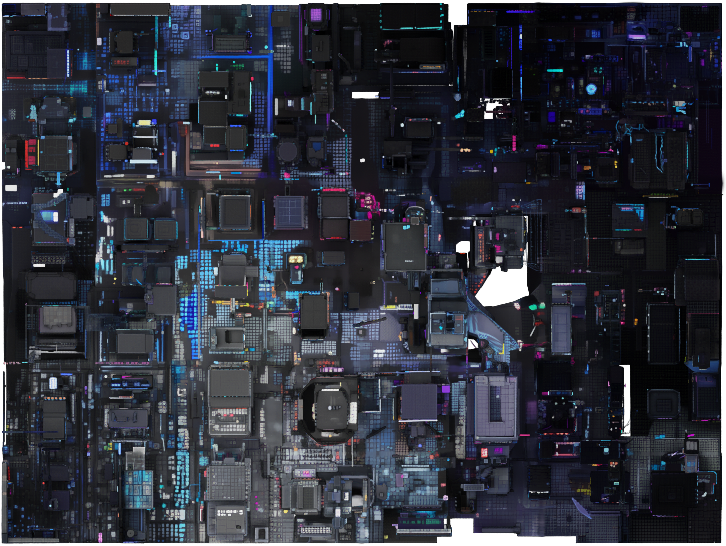} &
    \includegraphics[width=\imgw]{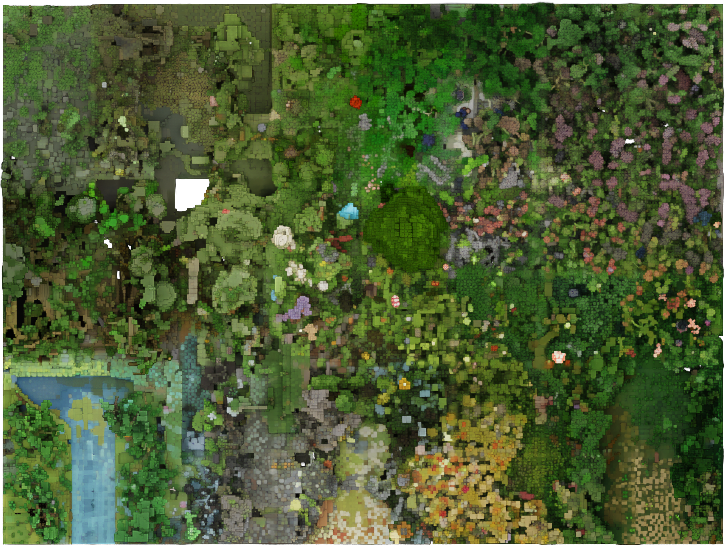}
    \\[6pt]
    &
    Desert & Cyberpubk & Forest
    \\
    \end{tabular}
    \caption{Top-down views of a generated 4x3x1 scene (not cherry-picked) using (a) an autoregressive method based on inpainting and (b) our method. Our method consistently shows better blending between tiles across different themes.}
    \label{fig:ablation_tiled_diffusion}
\end{figure}

\paragraph{Tiled Diffusion} A naive approach is to generate the world by stitching together multiple object generation results autoregressively, ensuring context alignment of neighboring chunks using inpainting. However, this leads to less coherent generation at the chunk edges, as shown in \autoref{fig:ablation_tiled_diffusion}.

\paragraph{Tiled Decoder}

\begin{figure}[t]
    \centering
    \includegraphics[width=\linewidth]{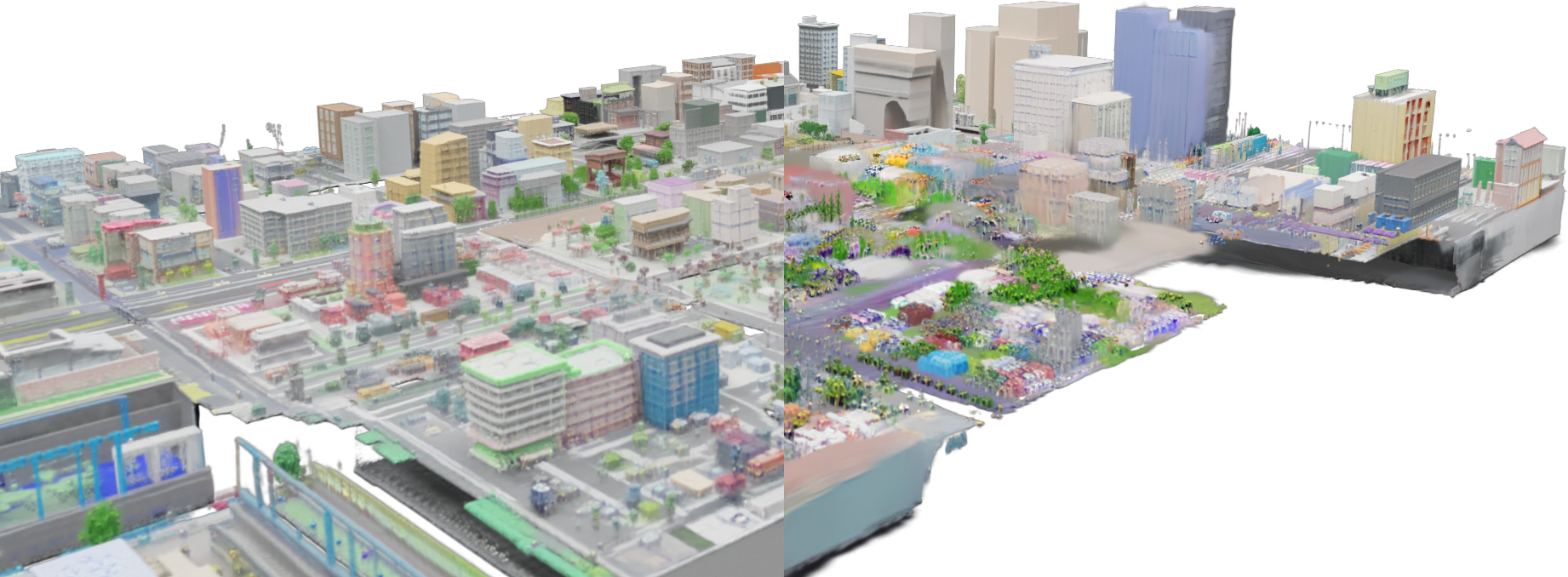}
    \begin{minipage}{0.49\linewidth}
        \centering
        (a) with tiled decoder
    \end{minipage}
    \begin{minipage}{0.49\linewidth}
        \centering
        (b) without tiled decoder
    \end{minipage}
    \caption{Comparison example with 3x2x1 city chunks for (a) decoding using our tiled decoding method and (b) decoding the entire generation at once. We observe severe artifacts when decoding without our tiled decoder.}
    \label{fig:ablation_tiled_decoder}
\end{figure}

We compare decoding using our tiled decoder with decoding the entire world at once in \autoref{fig:ablation_tiled_decoder}. Removing the tiled decoder shows artifacts in the generated Gaussian Splatting.

\paragraph{Blending}

\begin{figure}[t]
    \centering
    \newcommand{\lblw}{0.005\textwidth}   
    \newcommand{\imgw}{0.40\textwidth}   
    \begin{tabular}{%
    >{\centering\arraybackslash}m{\lblw}%
    >{\centering\arraybackslash}m{\imgw}%
    >{\centering\arraybackslash}m{\imgw}%
    }
    \rotatebox{90}{\scriptsize(a) Average Blending} &
    \includegraphics[width=\imgw]{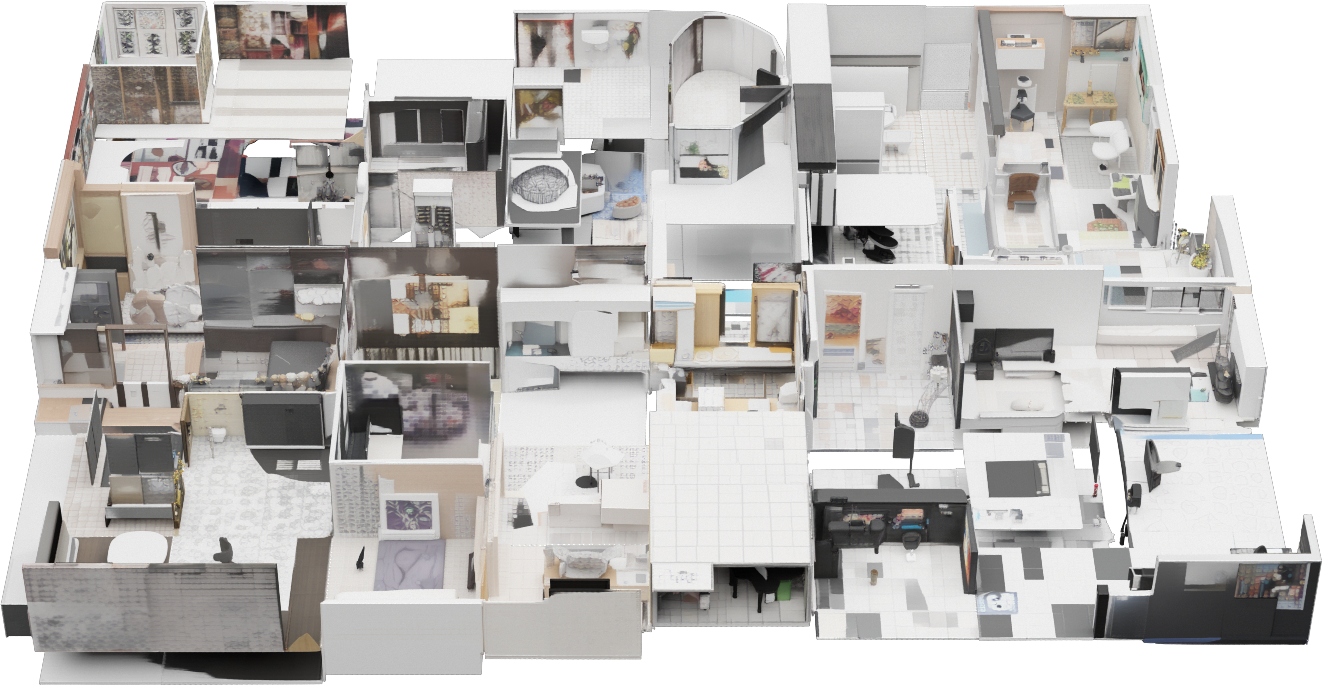} &
    \includegraphics[width=\imgw]{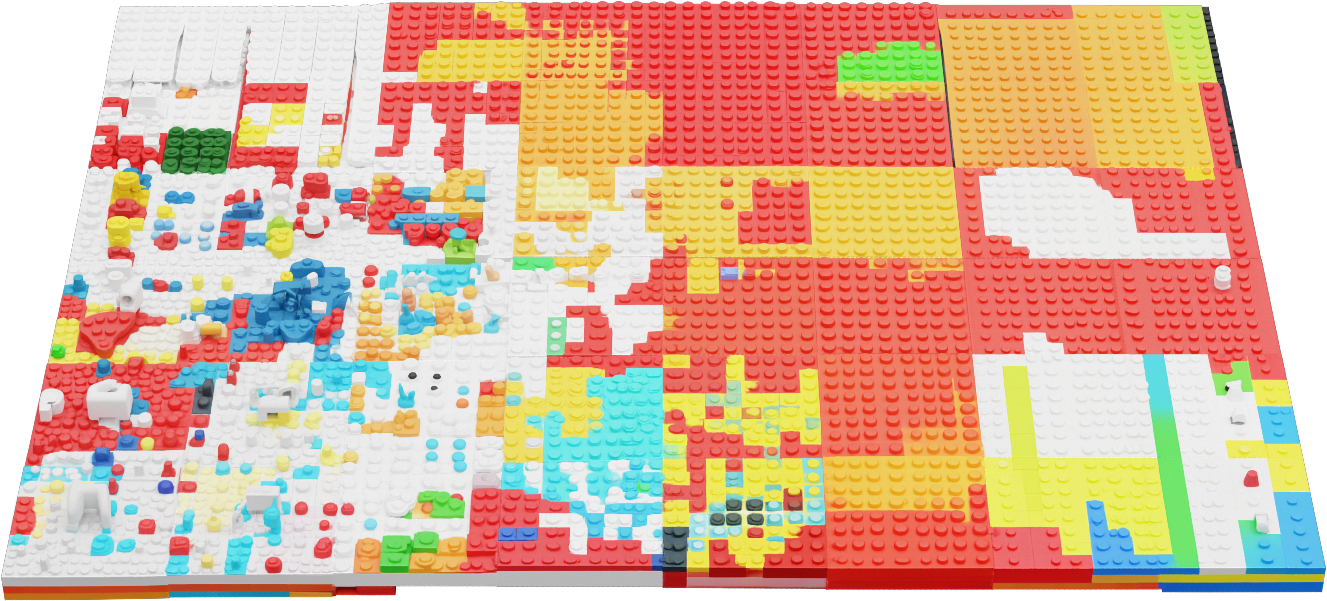}
    \\[4pt]
    \rotatebox{90}{\scriptsize(b) TRELLISWorld} &
    \includegraphics[width=\imgw]{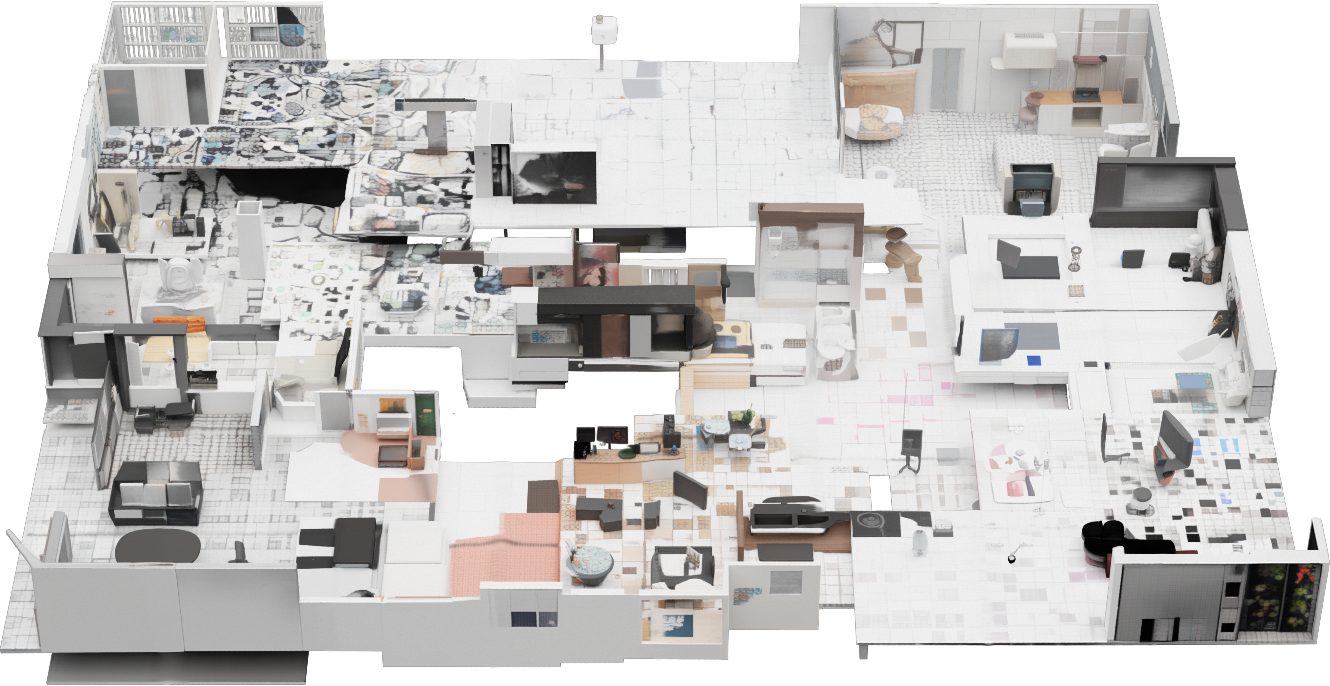} &
    \includegraphics[width=\imgw]{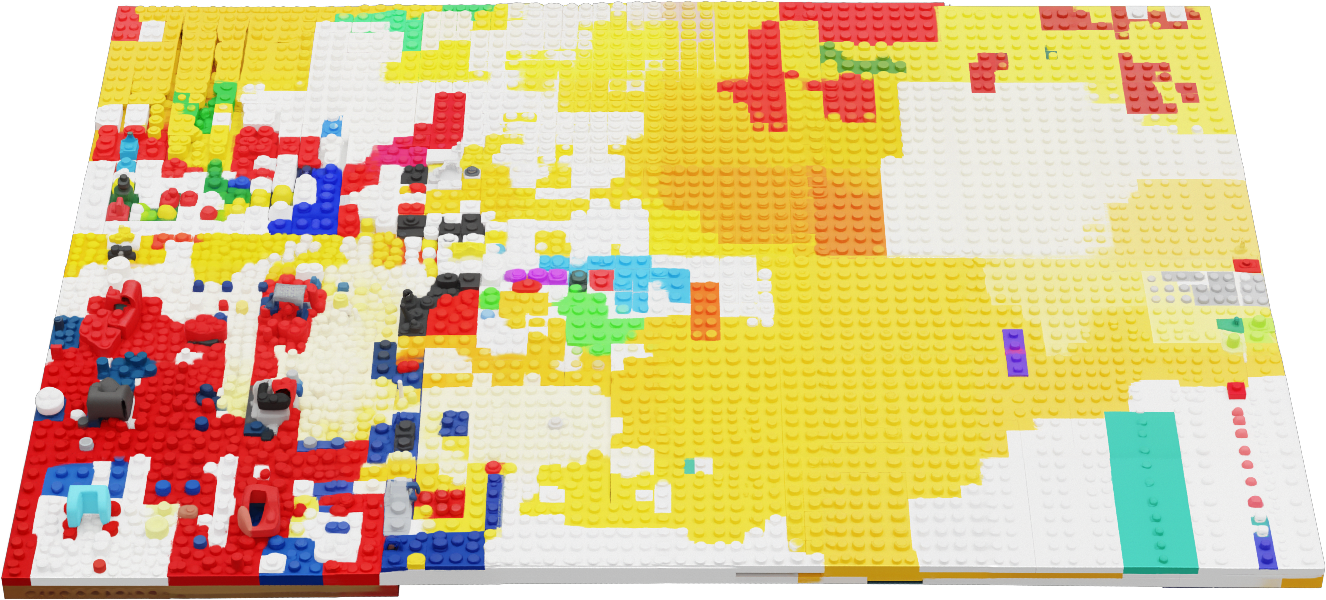}
    \\[6pt]
    &
    Room & Lego Tile
    \\
    \end{tabular}
    \caption{Comparison (not cherry-picked) showing the effectiveness of blending. (a) With average blending, the "room" example tends to generate walls around tile borders, and the "lego tile" example produces a colored edge along tile borders, which is undesirable. (b) Tile borders become less noticeable with blending.}
    \label{fig:ablation_blending}
\end{figure}

We replace our blending method with a simple averaging aggregation. This results in visible seams across tile borders, as shown in \autoref{fig:ablation_blending}.

\subsection{Qualitative Comparison}
\label{subsec:qualitative}

\begin{figure}[t]
    \centering
    \newcommand{\lblw}{0.005\textwidth}   
    \newcommand{\imgw}{0.25\textwidth}   
    \begin{tabular}{%
    >{\centering\arraybackslash}m{\lblw}%
    >{\centering\arraybackslash}m{\imgw}%
    >{\centering\arraybackslash}m{\imgw}%
    >{\centering\arraybackslash}m{\imgw}%
    }
    \rotatebox{90}{\scriptsize(a) SynCity} &
    \includegraphics[width=\imgw]{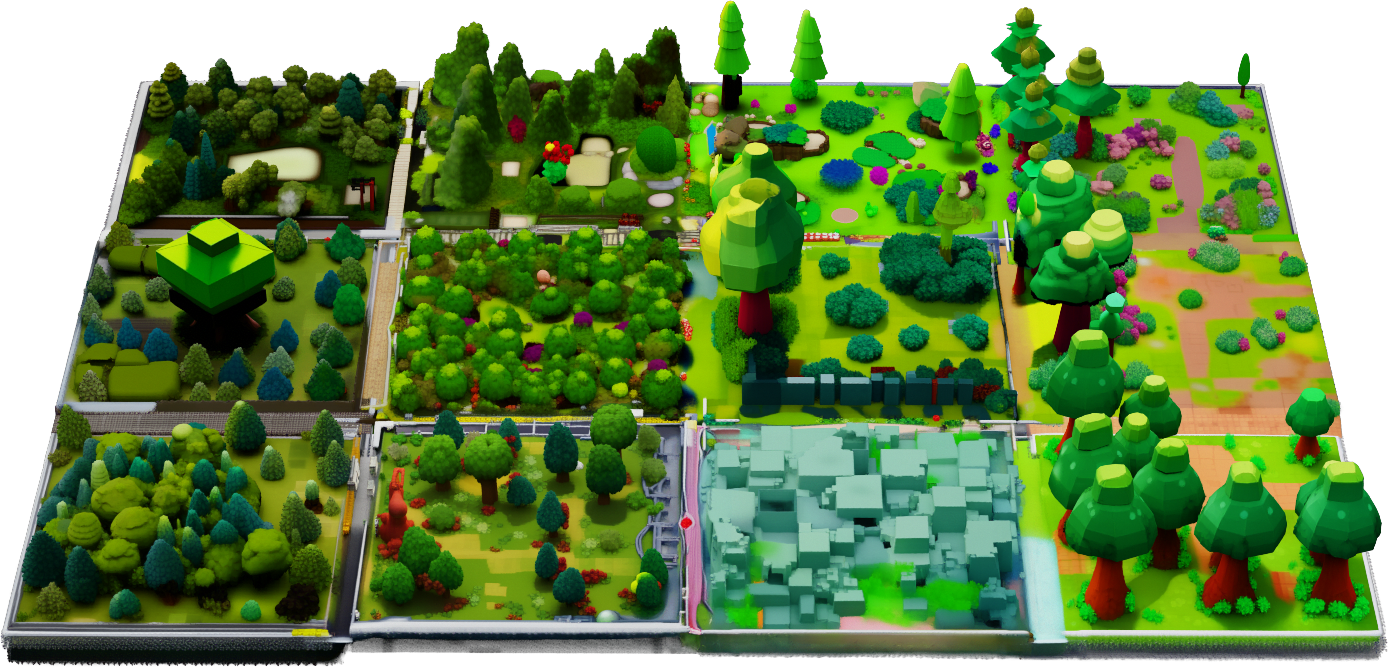} &
    \includegraphics[width=\imgw]{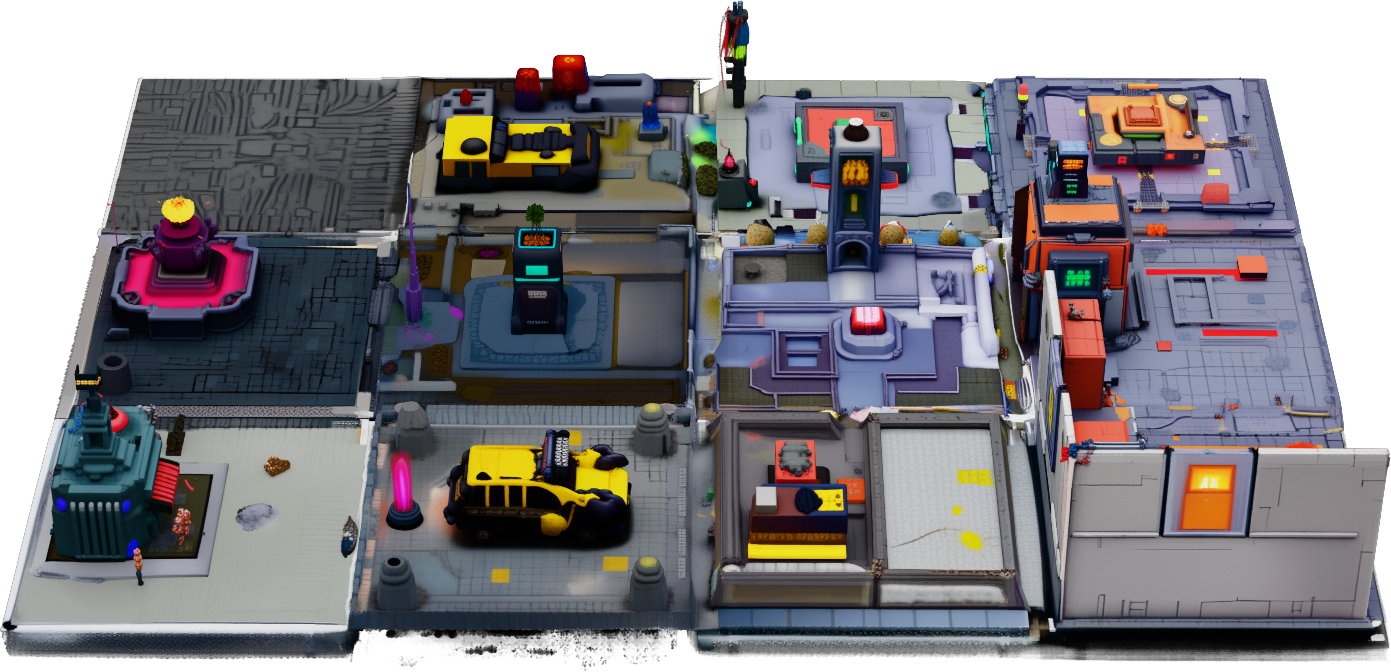} &
    \includegraphics[width=\imgw]{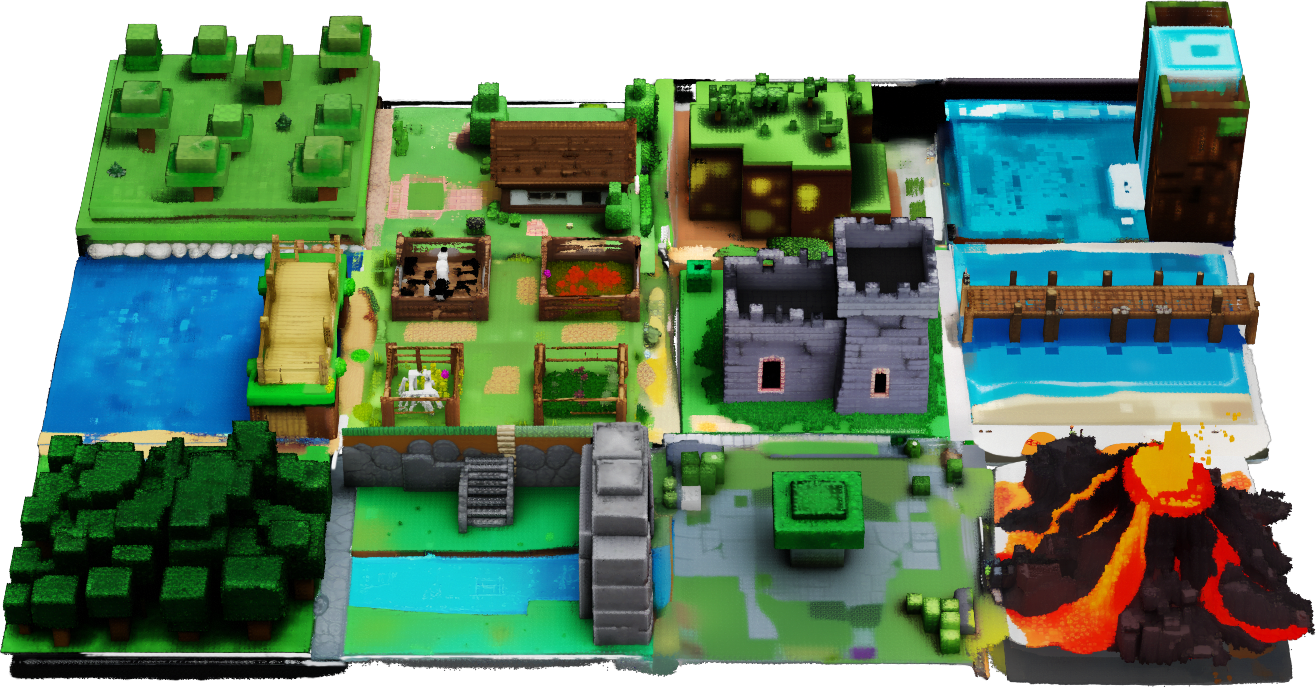}
    \\[4pt]
    \rotatebox{90}{\scriptsize(b) TRELLISWorld} &
    \includegraphics[width=\imgw]{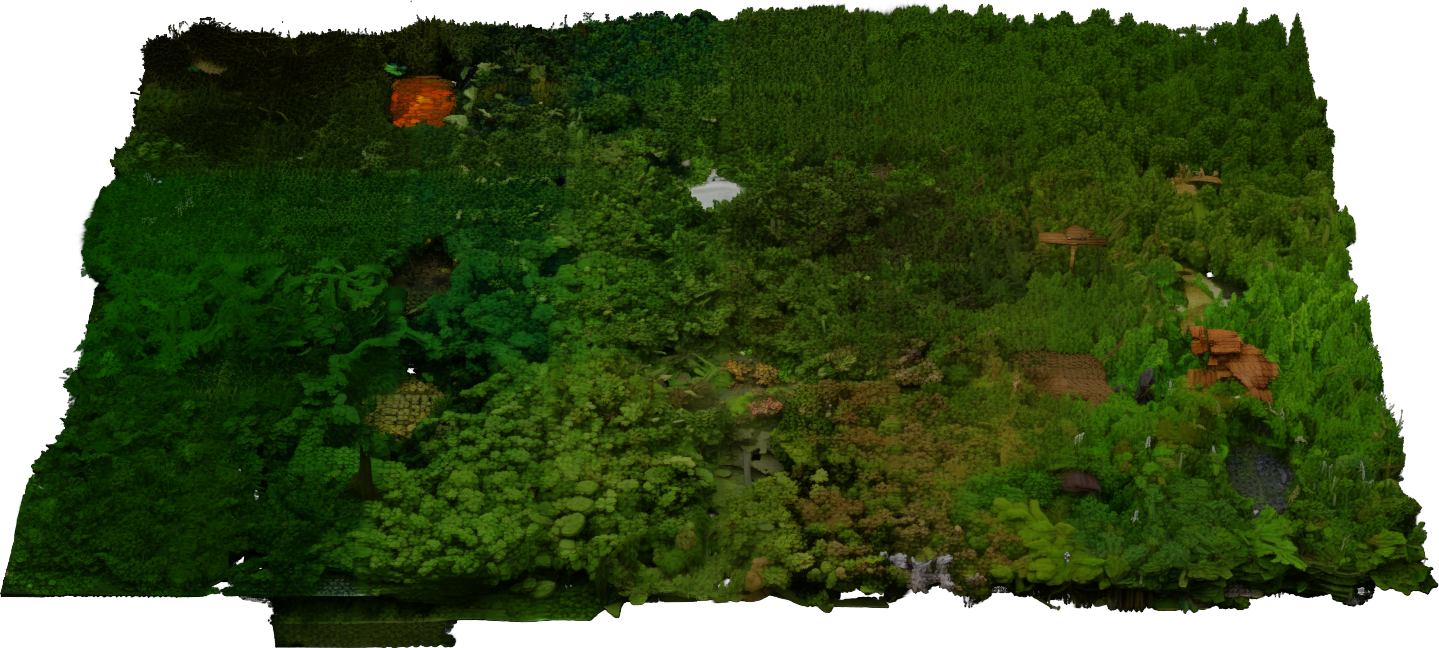} &
    \includegraphics[width=\imgw]{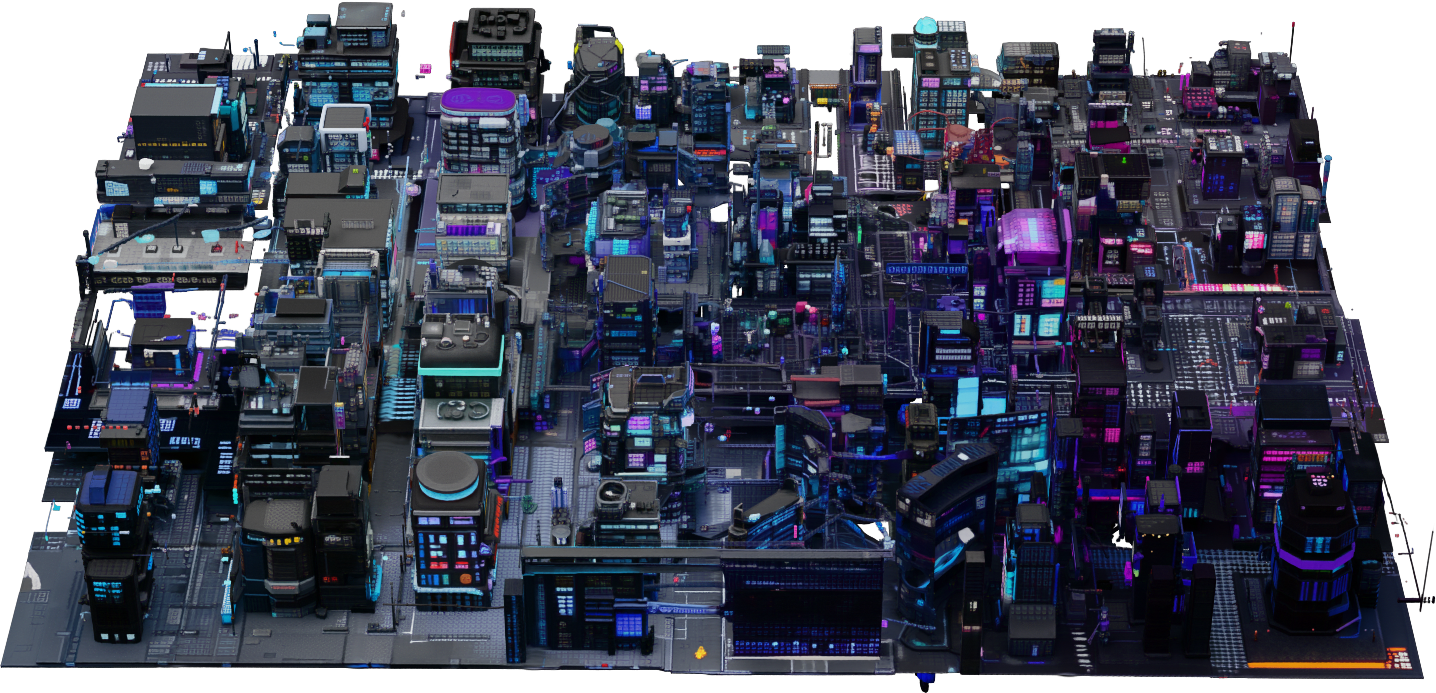} &
    \includegraphics[width=\imgw]{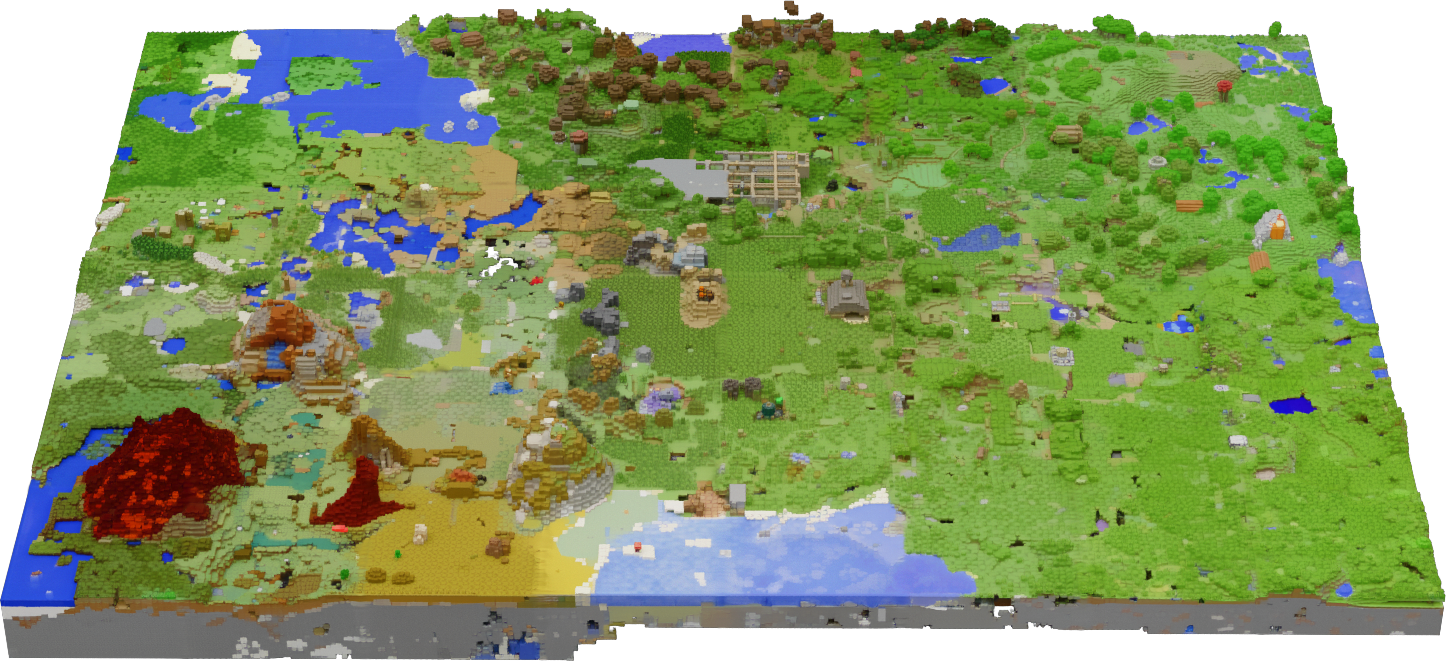}
    \\[6pt]
    &
    Forest & Cyberpubk & Minecraft
    \\
    \end{tabular}
    \caption{Qualitative comparison (not cherry-picked) between (a) SynCity \citep{syncity} and (b) TRELLISWorld (our method). All generations use a 4x3x1 chunk layout under the same Gaussian Splatting resolution. Our method demonstrates seamless blending between tiles, whereas chunk boundaries in SynCity are easily noticeable.
    }
    \label{fig:comparison}
\end{figure}

We compare our method with SynCity \citep{syncity}. \autoref{fig:comparison} showcases generation results across multiple prompts using the same 4x3x1 chunk layout. Our method tends to generate larger scenes and provides more natural blending across tiles compared to SynCity. For additional generation results of TRELLISWorld, see \autoref{fig:appendix_additional_1} and \autoref{fig:appendix_additional_2} in \autoref{sec:appendix}.

\begin{figure}[t]
    \centering
    \includegraphics[width=0.9\linewidth]{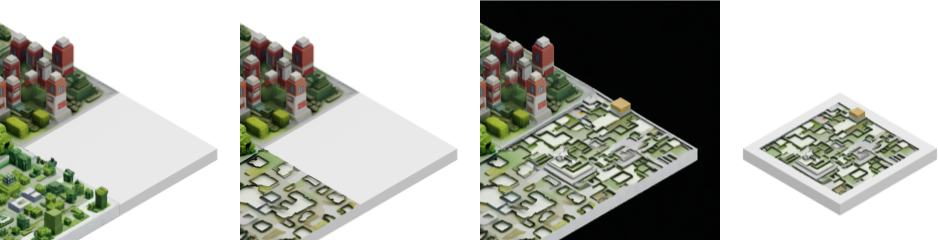}
    \begin{minipage}{0.24\linewidth}
        \centering
        (a) previous tiles
    \end{minipage}
    \begin{minipage}{0.24\linewidth}
        \centering
        (b) apply cropping
    \end{minipage}
    \begin{minipage}{0.24\linewidth}
        \centering
        (b) inpainted image
    \end{minipage}
    \begin{minipage}{0.24\linewidth}
        \centering
        (b) generated next tile
    \end{minipage}
    \caption{Limitations of world generation using autoregressive image inpainting methods described by SynCity \citep{syncity}. In SynCity, to prevent tall buildings from previous chunks from occluding the next chunk, the previously generated result in (a) is cropped to (b), producing an inpainted image with artifacts in (c). The object generator is then conditioned on image (c) to generate Gaussian Splatting (d), which contains inherited artifacts.}
    \label{fig:syncity_fail}
\end{figure}

Furthermore, our method is more robust compared to autoregressive methods based on image inpainting. For example, SynCity relies on heuristics such as cropping the 3D generation to avoid occlusion before applying 2D inpainting. Such heuristics are prone to failure when the 2D diffusion model mimics the cropped content, producing 3D chunks with artifacts, as shown in \autoref{fig:syncity_fail}.

\subsection{Quantitative Comparison}
\label{subsec:quantitative}

\paragraph{Perceptual Alignments}

\begin{table}[t]
    \centering
    \caption{Perceptual scores and CannyAvg error with standard deviation and confidence intervals across different methods. Our proposed method achieves the highest CLIP \citep{clip} score and lowest CannyAvg error, indicating better alignment with users' prompts as well as minimal seams between tiles.}
    \vspace{4pt}
    \label{tab:perceptual-alignments}
    \begin{tabular}{llll}
        \toprule
        \textbf{Method} & \textbf{CLIP Mean~$\uparrow$} & \textbf{CLIP STD} & \textbf{CLIP 95\%} \\
        \midrule
        SynCity & \textcolor{gray}{0.26020} & \textcolor{black}{0.02322} & [0.25726, 0.26314] \\
        inpaint baseline & \textcolor{darkgray}{0.26419} & \textcolor{black}{0.02534} & [0.26099, 0.26740] \\
        avg. blending & \textcolor{gray}{0.26203} & \textcolor{black}{0.02704} & [0.25861, 0.26545] \\
        TRELLISWorld & \textcolor{black}{0.26520} & \textcolor{black}{0.02496} & [0.26204, 0.26836] \\
        \midrule
        \textbf{} & \textbf{CannyAvg Mean~$\uparrow$} & \textbf{CannyAvg STD} & \textbf{CannyAvg 95\%} \\
        \midrule
        SynCity & \textcolor{gray}{7.81725} & \textcolor{black}{2.31850} & [6.98759, 8.64691] \\
        inpaint baseline & \textcolor{gray}{7.71797} & \textcolor{black}{5.63712} & [5.70075, 9.73519] \\
        avg. blending & \textcolor{darkgray}{6.55861} & \textcolor{black}{6.32146} & [4.29650, 8.82072] \\
        TRELLISWorld & \textcolor{black}{5.61331} & \textcolor{black}{5.49558} & [3.64674, 7.57987] \\
        \bottomrule
    \end{tabular}
\end{table}

To compare perceptual alignments with SynCity using the CLIP score based on \textit{clip-vit-base-patch32} model \citep{clip}, we uniformly rendered $18$ views at close distance from $15$ generation results, each of size 4x3x1 across diverse prompts, visualized in \autoref{fig:appendix_metrics_clip}. We adopt the same set of prompts (e.g., "city", "medieval", "desert", ...) for both methods following the procedure detailed in \autoref{appendix:prompt}. The result in \autoref{tab:perceptual-alignments} only shows marginal improvements, as the CLIP score does not directly measure blending quality across tiles.

\paragraph{Seam Visibility}

The main contribution of our method lies in reducing seams between incoherent object generations. To measure the visibility of the seams, we first rendered the scene using orthographic projection and then calculated a binary map using the Canny edge detection operator. We use the average intensity of this binary map to define the seam visibility metric CannyAvg. You can find a detailed explanation of CannyAvg in \autoref{appendix:metrics_cannyavg}.

\begin{figure}[t]
    \centering
    \begin{minipage}{0.48\textwidth}
    \begin{tikzpicture}
        \begin{axis}[
            title={(a) CannyAvg by Different Strides},
            width=6.5cm,
            height=3.5cm,
            xlabel={Stride $s$ ($S=16$)},
            ylabel={CannyAvg},
            grid=both,
            legend style={at={(0.03,0.97)},anchor=north west,font=\scriptsize},
            legend image post style={
                scale=0.5,
            },
            scaled x ticks = false,
            scaled y ticks = false,
            x tick label style={/pgf/number format/fixed},
            y tick label style={/pgf/number format/fixed},
            x dir=reverse,
        ]
        \addplot[
            color=red!40,
            thick,
            mark=square*,
            mark options={fill=red!80, draw=red!80},
        ]
        coordinates {
            (16,5.631237030029297)
            (12,4.068436622619629)
            (8,2.596750259399414)
            (6,3.0912303924560547)
            (4,2.1401262283325195)
            (2,2.0943260192871094)
        };
        \end{axis}
    \end{tikzpicture}
    \end{minipage}
    \begin{minipage}{0.48\textwidth}
    \begin{tikzpicture}
        \begin{axis}[
            title={(b) CannyAvg by Computation},
            width=6.5cm,
            height=3.5cm,
            xlabel={Avg. Sec./Scene},
            grid=both,
            legend style={at={(0.03,0.97)},anchor=north west,font=\scriptsize},
            legend image post style={
                scale=0.5,
            },
            scaled x ticks = false,
            scaled y ticks = false,
            x tick label style={/pgf/number format/fixed},
            y tick label style={/pgf/number format/fixed},
        ]
        \addplot[
            color=red!40,
            thick,
            mark=square*,
            mark options={fill=red!80, draw=red!80},
        ]
        coordinates {
            (336.90,5.631237030029297)
            (561.27,4.068436622619629)
            (885.22,2.596750259399414)
            (1566.32,3.0912303924560547)
            (3260.95,2.1401262283325195)
            (11676.29,2.0943260192871094)
        };
        \end{axis}
    \end{tikzpicture}
    \end{minipage}
    \caption{CannyAvg measuring seams in a scene using variable stride $s$. All experiments are conducted on a 4x3x1 chunk layout with a fixed prompt. Decreasing stride increases scene quality, with $s=8$ being an optimal trade-off between computational time and scene quality.}
    \label{fig:cannyavg-cost-tiles}
\end{figure}
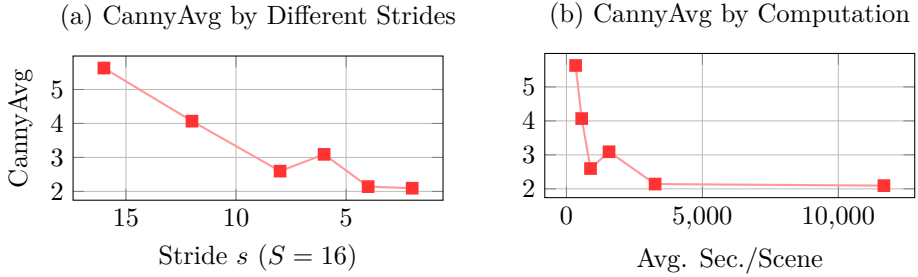

\paragraph{Computational Cost}

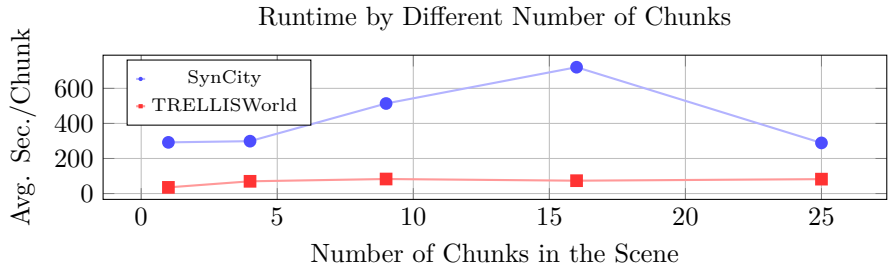
\begin{figure}[t]
    \centering
    \begin{tikzpicture}
        \begin{axis}[
            title={Runtime by Different Number of Chunks},
            width=12cm,
            height=3.5cm,
            xlabel={Number of Chunks in the Scene},
            ylabel={Avg. Sec./Chunk},
            grid=both,
            legend style={at={(0.03,0.97)},anchor=north west,font=\scriptsize},
            legend image post style={
                scale=0.2,
            },
        ]
        \addplot[
            color=blue!30,
            thick,
            mark=*,
            mark options={fill=blue!70, draw=blue!70},
        ]
        coordinates {
            (1,292.00)
            (4,298.75)
            (9,513.67)
            (16,720.69)
            (25,288.84)
        };
        \addlegendentry{SynCity}
        \addplot[
            color=red!40,
            thick,
            mark=square*,
            mark options={fill=red!80, draw=red!80},
        ]
        coordinates {
            (1,35.18)
            (4,69.85)
            (9,82.89)
            (16,73.34)
            (25,82.16)
        };
        \addlegendentry{TRELLISWorld}

        \end{axis}
    \end{tikzpicture}
    \caption{Computational cost required to generate a scene of variable scene sizes. All experiments are conducted on square chunk layouts from 1x1x1 to 5x5x1 with a fixed prompt. TRELLISWorld takes less time to generate a scene for all scene sizes.}
    \label{fig:computational-cost-tiles}
\end{figure}

\begin{figure}[t]
    \centering
    \begin{minipage}{0.48\textwidth}
    \begin{tikzpicture}
        \begin{axis}[
            title={(a) Runtime by Different Strides},
            width=6.5cm,
            height=3.5cm,
            xlabel={Stride $s$ ($S=16$)},
            ylabel={Avg. Sec./Scene},
            grid=both,
            legend style={at={(0.03,0.97)},anchor=north west,font=\scriptsize},
            legend image post style={
                scale=0.5,
            },
            scaled x ticks = false,
            scaled y ticks = false,
            x tick label style={/pgf/number format/fixed},
            y tick label style={/pgf/number format/fixed},
            x dir=reverse,
        ]
        \addplot[
            color=red!40,
            thick,
            mark=square*,
            mark options={fill=red!80, draw=red!80},
        ]
        coordinates {
            (16,336.90)
            (12,561.27)
            (8,885.22)
            (6,1566.32)
            (4,3260.95)
            (2,11676.29)
        };
        \end{axis}
    \end{tikzpicture}
    \end{minipage}
    \begin{minipage}{0.48\textwidth}
    \begin{tikzpicture}
        \begin{axis}[
            title={(b) Runtime by Different Tiles},
            width=6.5cm,
            height=3.5cm,
            xlabel={Number of Tiles},
            grid=both,
            legend style={at={(0.03,0.97)},anchor=north west,font=\scriptsize},
            legend image post style={
                scale=0.5,
            },
            scaled x ticks = false,
            scaled y ticks = false,
            x tick label style={/pgf/number format/fixed},
            y tick label style={/pgf/number format/fixed},
        ]
        \addplot[
            color=red!40,
            thick,
            mark=square*,
            mark options={fill=red!80, draw=red!80},
        ]
        coordinates {
            (12,336.90)
            (20,561.27)
            (35,885.22)
            (63,1566.32)
            (117,3260.95)
            (425,11676.29)
        };
        \end{axis}
    \end{tikzpicture}
    \end{minipage}
    \caption{Computational cost required to generate a scene for variable stride $s$. All experiments are conducted on the 4x3x1 chunk layouts with a fixed prompt. Decreasing $s$ reciprocally increases the number of tiles, but the time to generate a tile remains constant ($27.21$ seconds per tile on average)}
    \label{fig:computational-cost-stride}
\end{figure}
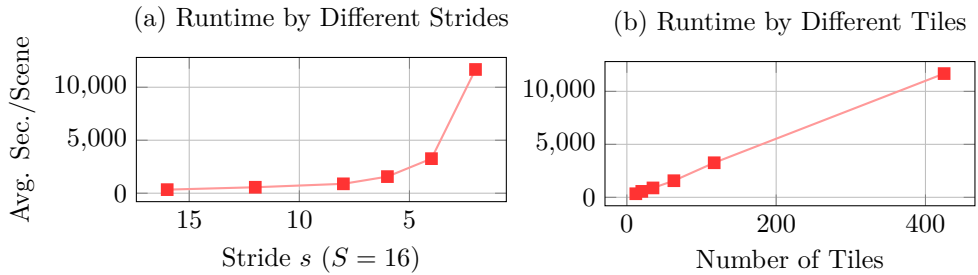

Without any optimization, TRELLISWorld takes $77.96$ seconds on average to generate a chunk, which is a $5.80\times$ speedup over SynCity \citep{syncity} ($452.04$ seconds). TRELLISWorld also significantly reduces memory cost. During comparison to SynCity, we found that SynCity is unable to run on a single NVIDIA GeForce RTX 4080 (16GB). Therefore, we resort all SynCity experiments to a single NVIDIA GeForce RTX 4090 (48GB), a better GPU on all metrics, giving SynCity an unfair advantage. \autoref{fig:computational-cost-tiles} and \autoref{fig:computational-cost-stride} show a linear increase in runtime as the number of chunks or tiles increases. Moreover, because our method does not rely on autoregressive inference, larger scenes can potentially be parallelized across multiple GPUs for further speedup.

\section{Applications}

\begin{figure}[t]
    \centering
    \includegraphics[width=0.24\linewidth]{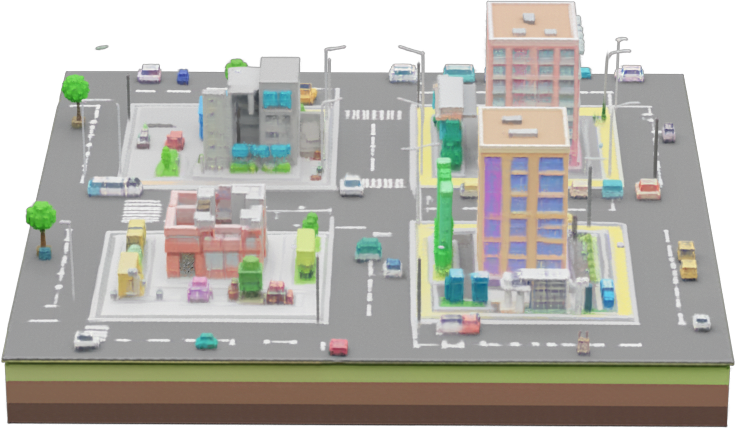}
    \includegraphics[width=0.24\linewidth]{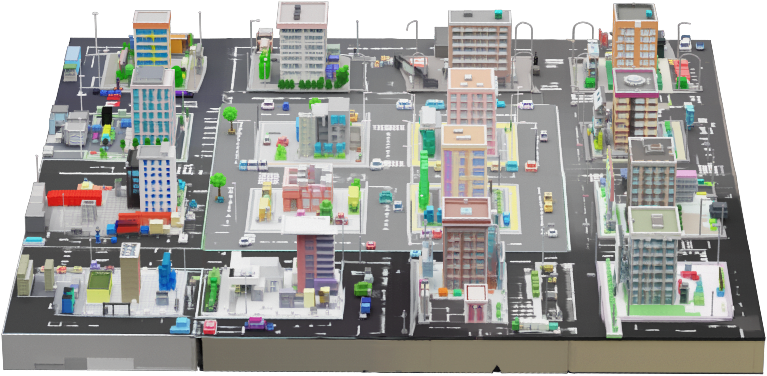}
    \includegraphics[width=0.24\linewidth]{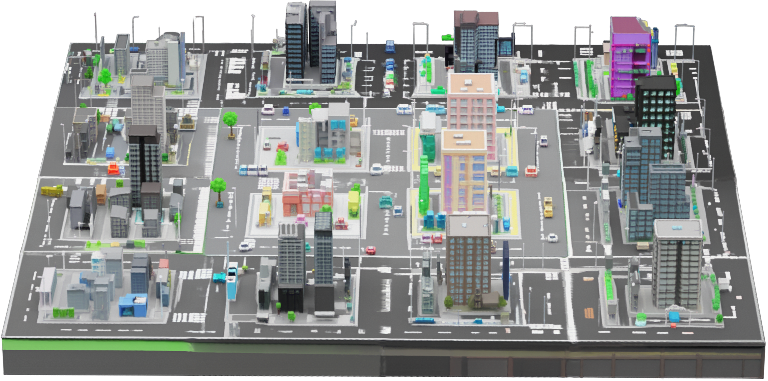}
    \includegraphics[width=0.24\linewidth]{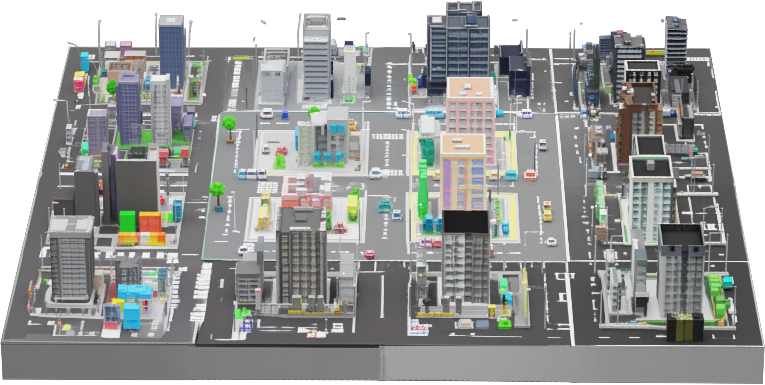}
    \caption{City scene expansion results (not cherry-picked) using TRELLISWorld. Given the leftmost 1x1x1 chunk as input, the model generates a 3x3x1 extended scene. Three diverse outputs are shown to the right, demonstrating variations.}
    \label{fig:application_inpaint}
\end{figure}

\paragraph{Editing or Expanding Existing Worlds}  
Our method can expand already-generated scenes by initializing the noise with parts of the ground truth, as shown in \autoref{fig:application_inpaint} and detailed in \autoref{appendix:inpaint}.

\begin{figure}[b]
    \centering
    \begin{minipage}[b]{0.48\linewidth}
        \centering
        \includegraphics[width=\linewidth, trim=0 0 0 50, clip]{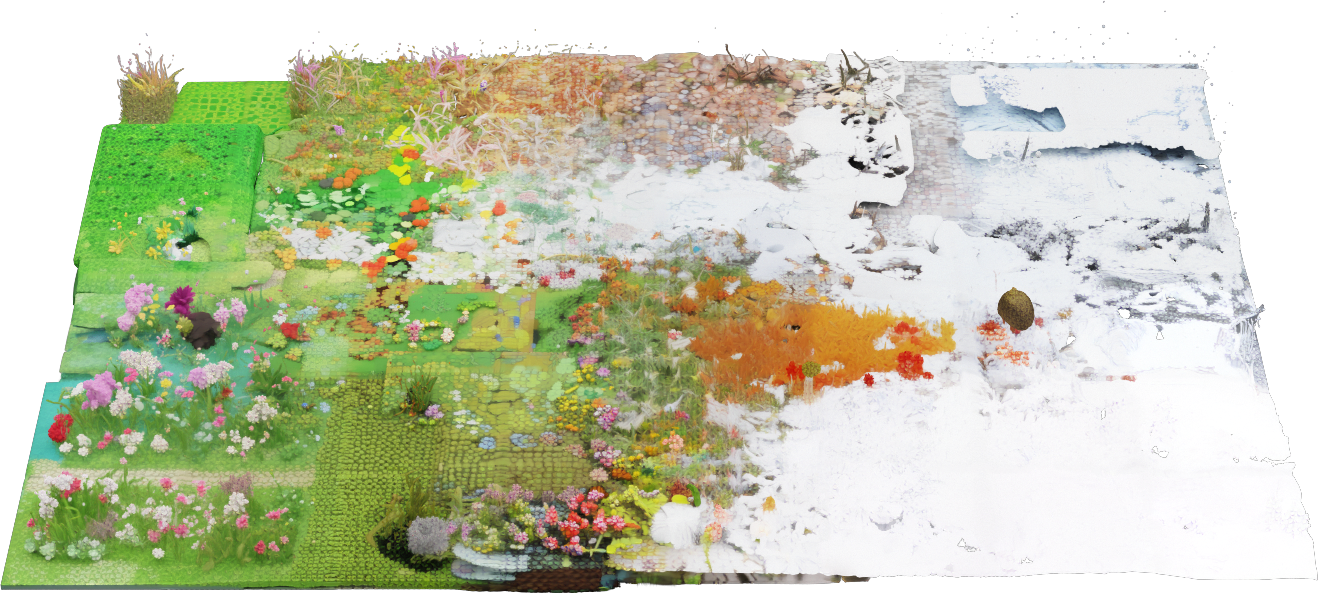}
        \caption{Generation result (not cherry-picked) showing a smooth and natural transition from "Spring forest tile... blooming flowers..." (bottom-left) to "Winter ice lake... skating marks..." (top-right).}
        \label{fig:application_transition}
        {}
    \end{minipage}
    \hfill
    \begin{minipage}[b]{0.48\linewidth}
        \centering
        \begin{minipage}[b]{0.48\linewidth}
            \centering
            \includegraphics[width=\linewidth, trim=0 120 0 0, clip]{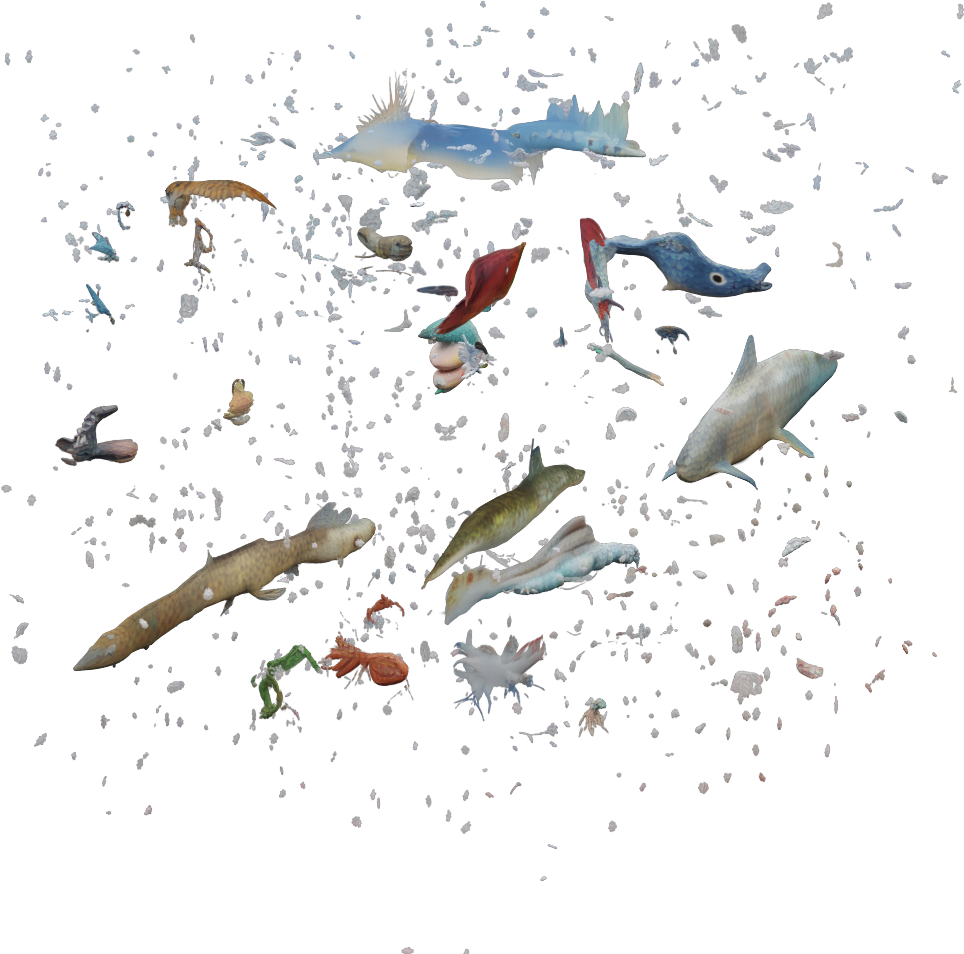}
            {\small (a) Group of Fish (2x2x2 chunk)}
        \end{minipage}
        \hfill
        \begin{minipage}[b]{0.48\linewidth}
            \centering
            \begin{minipage}[b]{0.48\linewidth}
                \centering
                \includegraphics[width=\linewidth]{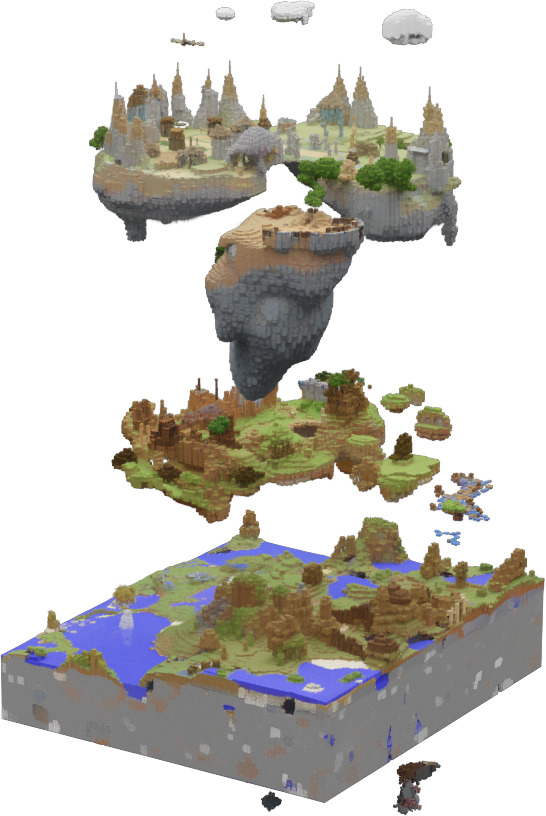}
            \end{minipage}
            \hfill
            \begin{minipage}[b]{0.48\linewidth}
                \centering
                \includegraphics[width=\linewidth]{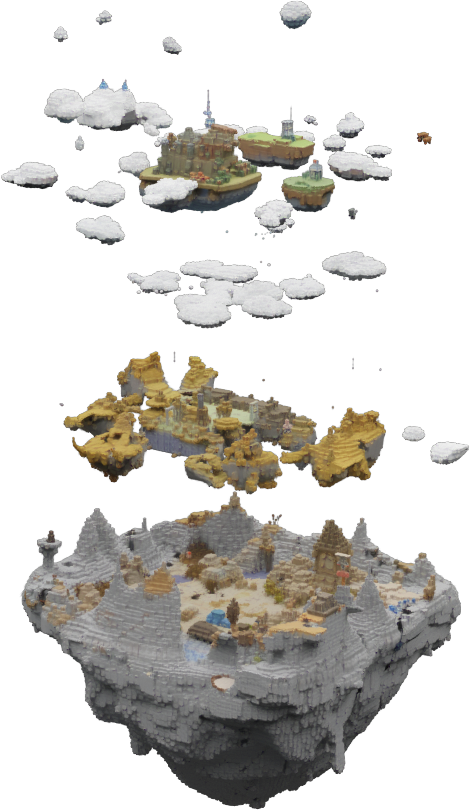}
            \end{minipage}
            {\small (b) Castle in the Sky (1x1x2 chunk)}
        \end{minipage}
        \caption{Examples of three-dimensional tiling. The left scene is created using a uniform prompt: "A group of fish swimming in the air". The right scenes are two variations created using area-specific prompting.}
        \label{fig:application_3d}
    \end{minipage}
\end{figure}

\paragraph{Area-Specific Prompting}  
Our method allows users to specify different prompts at each location. We demonstrate this capability in \autoref{fig:application_transition}. See \autoref{appendix:prompt} for details on creating area-specific prompts.

\paragraph{Three-Dimensional Tiling}  
While most macro-structures on Earth are constrained to two-dimensional surfaces, our method naturally generalizes to the generation of three-dimensional macro-structures, such as a group of fish, or 3D blending using the area-specific prompting technique described above, as shown in \autoref{fig:application_3d}. To our knowledge, no existing method offers this level of flexibility.

\section{Limitations}

While our method successfully generates coherent scenes without training, it presents several limitations for future investigations:

\paragraph{Dependence on Base Models}  
As a training-free approach, our method is inherently constrained by the capabilities of the underlying base models. In particular, the performance of TRELLIS directly limits both the visual fidelity and the efficiency of our scene generation pipeline.

\paragraph{Object-Level Separation}  
To ensure global scene coherence, our method performs generation in a single batch. As a result, it lacks the ability to disentangle individual objects post-generation.

\section{Conclusion}

We presented TRELLISWorld, a training-free framework for text-driven 3D scene generation that composes large-scale environments by repurposing object-level diffusion models through a tiled denoising formulation. By leveraging spatial overlap and cosine-weighted blending, our method enables semantically coherent, scalable, and editable 3D world synthesis without retraining. Experiments demonstrate that TRELLISWorld outperforms existing autoregressive approaches in both visual coherence and computational efficiency, while supporting flexible applications such as localized prompting. Our results establish a simple yet extensible foundation for general-purpose language-guided 3D scene construction, bridging the gap between object-level priors and world-scale generation.

\section*{Acknowledgement}
We sincerely thank Zekai Gu for sharing resources, and anonymous reviewers for their valuable feedback.

\bibliography{iclr2026_conference}
\bibliographystyle{iclr2026_conference}

\appendix

\section{Appendix}
\label{sec:appendix}

\subsection{Text Prompt}
\label{appendix:prompt}

Our method gives the user explicit control of the 3D prompts used in generation. Below is an example of a 3D prompt we use to generate our 4x3x1 city theme:

\begin{lstlisting}[]
prompt = [[
    ["Dense low-rise residential block with small shops and narrow streets, square tile"],
    ["Cluster of mid-rise apartments with pastel facades and tree-lined sidewalks, square tile"],
    ["Modern commercial zone with glass offices, cafes, and public seating, square tile"],
], [
    ["Mixed-use area with offices, apartments, and green courtyards, square tile"],
    ["Urban block with mid-rise towers, parking lots, and small plazas, square tile"],
    ["Dense retail and commercial buildings near busy intersection, square tile"],
], [
    ["Residential zone with consistent low-rise buildings and local shops, square tile"],
    ["Compact city block with modern mid-rises and organized street grid, square tile"],
    ["Edge of city with fewer high-rises and more greenery, square tile"],
], [
    ["Park extension with dense trees and a water feature, square tile"],
    ["Community recreation area with playgrounds and open lawns, square tile"],
    ["Park transition with scattered cultural buildings and trees, square tile"],
]]
\end{lstlisting}

For other themes (``city'', ``medieval'', ``desert'', ``cyberpunk'', ``ancient Rome'', ``minecraft'', ``forest'', ``ocean'', ``winter'', ``lego'', ``park'', ``amusement park'', ``airport'', ``college'', ``room''), we ask LLMs to generate similar prompts using in-context learning from the city prompt above. For example, we generated our medieval prompt by asking "[IN\_CONTEXT\_PROMPT] Above is a prompt for creating a large city block with parks, please give a prompt for 'a medieval tile with farmland fields and cottages' following the same format."

As shown above, the input prompt is formatted as a 3D tensor. Therefore, prompts in spatial proximity will remain close in the generated scene. We do not force the user to provide a tensor that has the exact shape of our tiles, as this may limit flexibility and ease of use. Instead, we treat the input prompt tensor as a tensor spanning the entire scene. Specifically, given a tile location, we sample the nearest prompt in the prompt tensor. This way, users can provide a coarser prompt tensor.

\subsection{Inpaint}
\label{appendix:inpaint}

Our method can fill missing regions or extend a user-provided chunks. We use RePaint \citep{inpaint} with a Gaussian-blurred mask to partially preserve the edges of the input chunks. This encourages smoother transitions between the generated and existing contents.

\subsection{Metrics}

\subsubsection{CLIP} 
\begin{figure}[t]
  \centering
  \begin{minipage}[c][2.2cm][c]{0.06\textwidth}
    \centering
    \rotatebox[origin=c]{90}{SynCity}
  \end{minipage}%
  \begin{minipage}[c]{0.92\textwidth}
    \includegraphics[width=0.16\linewidth]{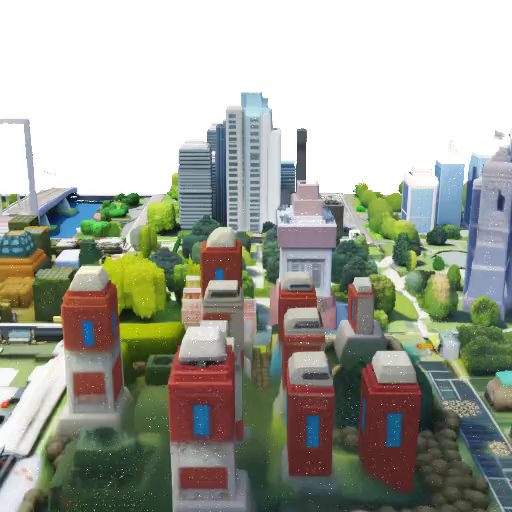}\hfill
    \includegraphics[width=0.16\linewidth]{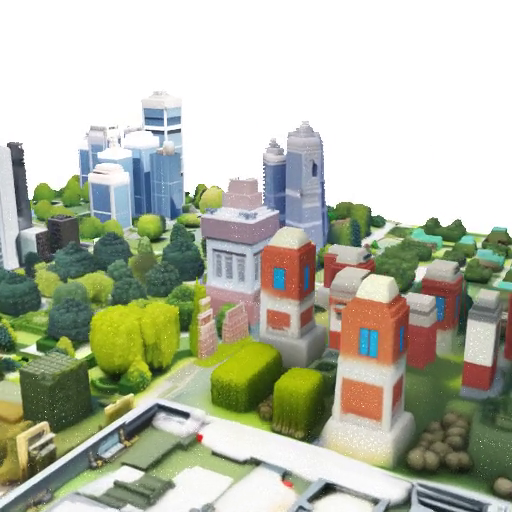}\hfill
    \includegraphics[width=0.16\linewidth]{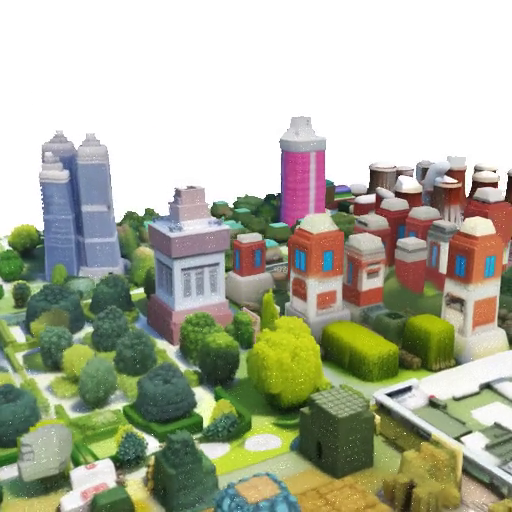}\hfill
    \includegraphics[width=0.16\linewidth]{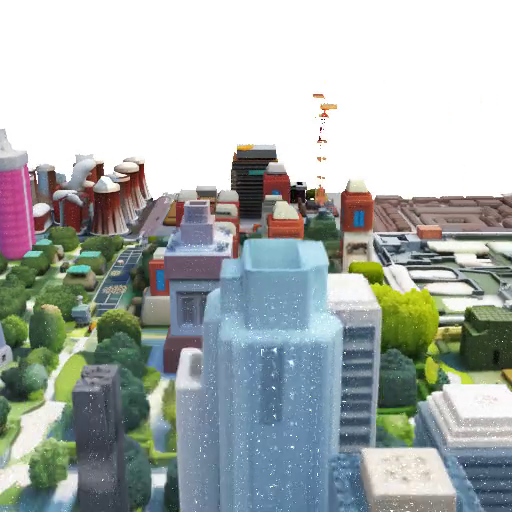}\hfill
    \includegraphics[width=0.16\linewidth]{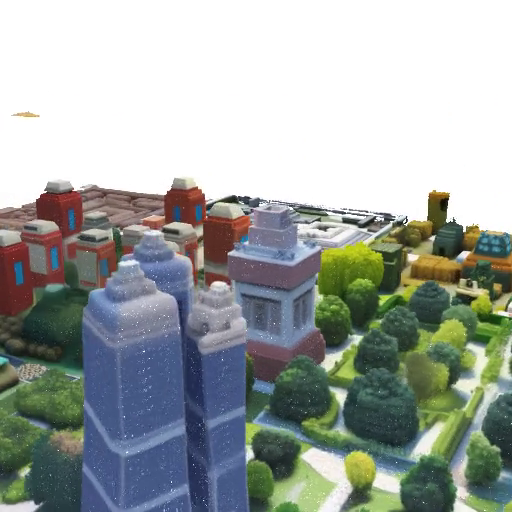}\hfill
    \includegraphics[width=0.16\linewidth]{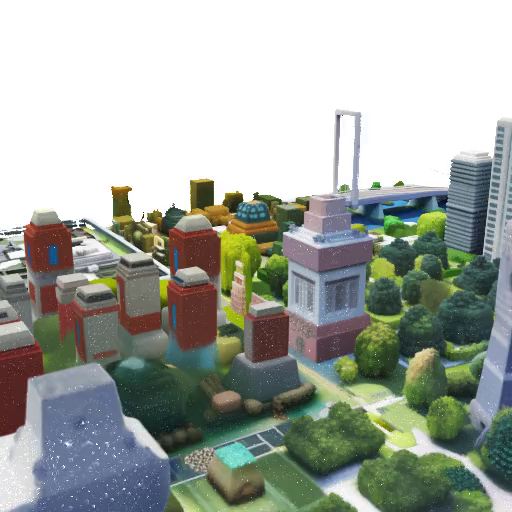}
  \end{minipage}
  \begin{minipage}[c][2.2cm][c]{0.06\textwidth}
    \centering
    \rotatebox[origin=c]{90}{TRELLISWorld}
  \end{minipage}%
  \begin{minipage}[c]{0.92\textwidth}
    \includegraphics[width=0.16\linewidth]{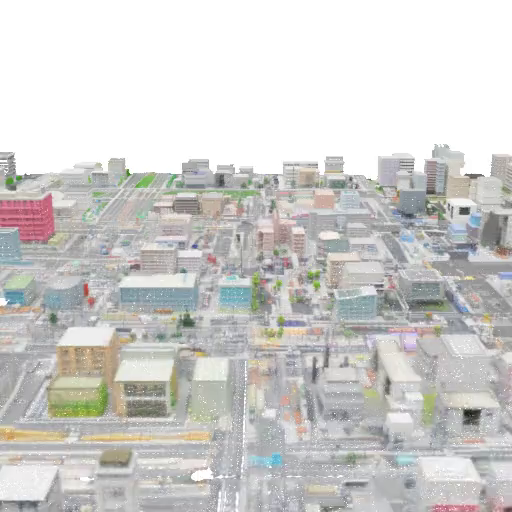}\hfill
    \includegraphics[width=0.16\linewidth]{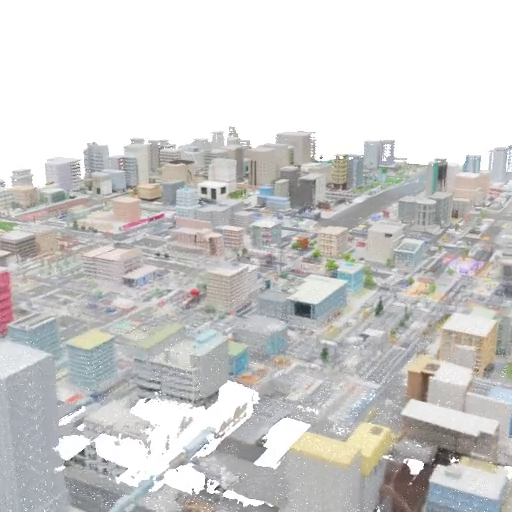}\hfill
    \includegraphics[width=0.16\linewidth]{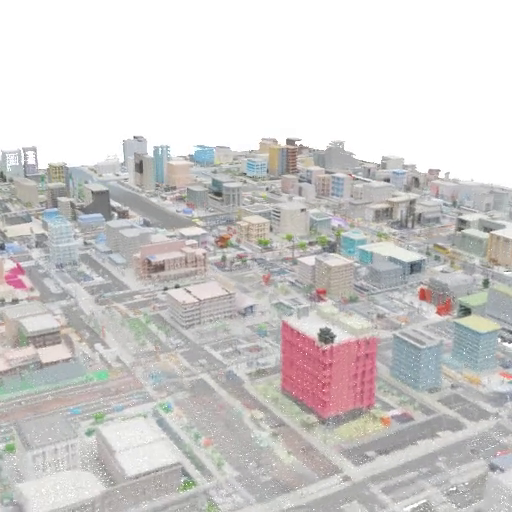}\hfill
    \includegraphics[width=0.16\linewidth]{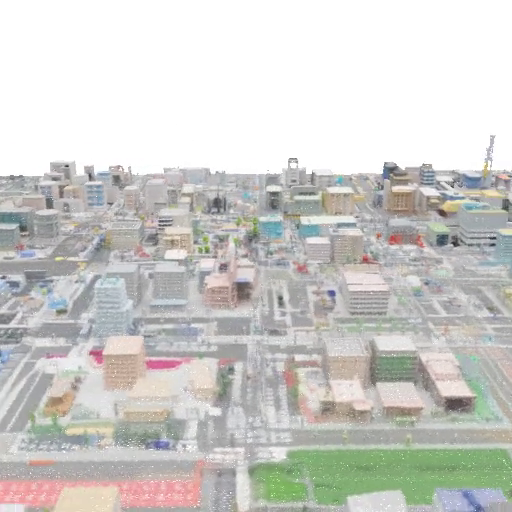}\hfill
    \includegraphics[width=0.16\linewidth]{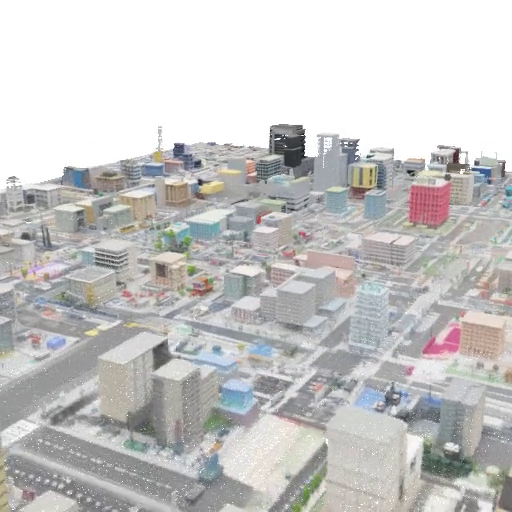}\hfill
    \includegraphics[width=0.16\linewidth]{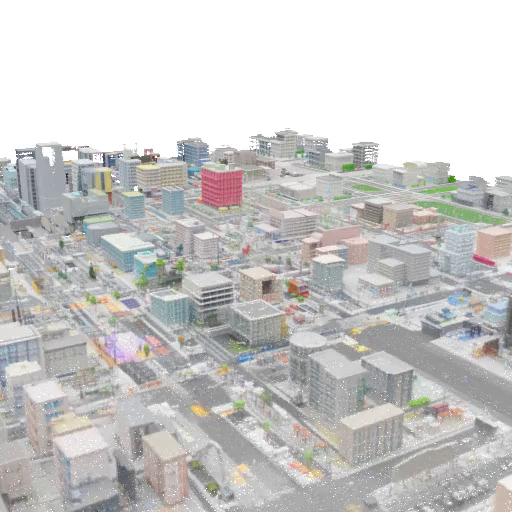}
  \end{minipage}

\caption{Sampled views for CLIP similarity calculation.}
\label{fig:appendix_metrics_clip}
\end{figure}

We measured CLIP similarity using $18$ views per generated scene. \autoref{fig:appendix_metrics_clip} shows $6$ different views from SynCity and TRELLISWorld respectively. We used the mega-prompt (e.g., 'a medieval tile with farmland fields and cottages') instead of the object-level prompt for CLIP text input, appending ", brid eye view" and removing all auxiliary modifications such as ", on top of a base" and ", square tile" used by SynCity and TRELLISWorld to better reflect the actual content in the images.

\subsubsection{CannyAvg}
\label{appendix:metrics_cannyavg} 
\begin{figure}[t]
    \centering
    \newcommand{\lblw}{0.005\textwidth}   
    \newcommand{\imgw}{0.21\textwidth}   
    \begin{tabular}{%
    >{\centering\arraybackslash}m{\lblw}%
    >{\centering\arraybackslash}m{\imgw}%
    >{\centering\arraybackslash}m{\imgw}%
    >{\centering\arraybackslash}m{\imgw}%
    >{\centering\arraybackslash}m{\imgw}%
    }
    \rotatebox{90}{s=2} &
    \includegraphics[width=\imgw]{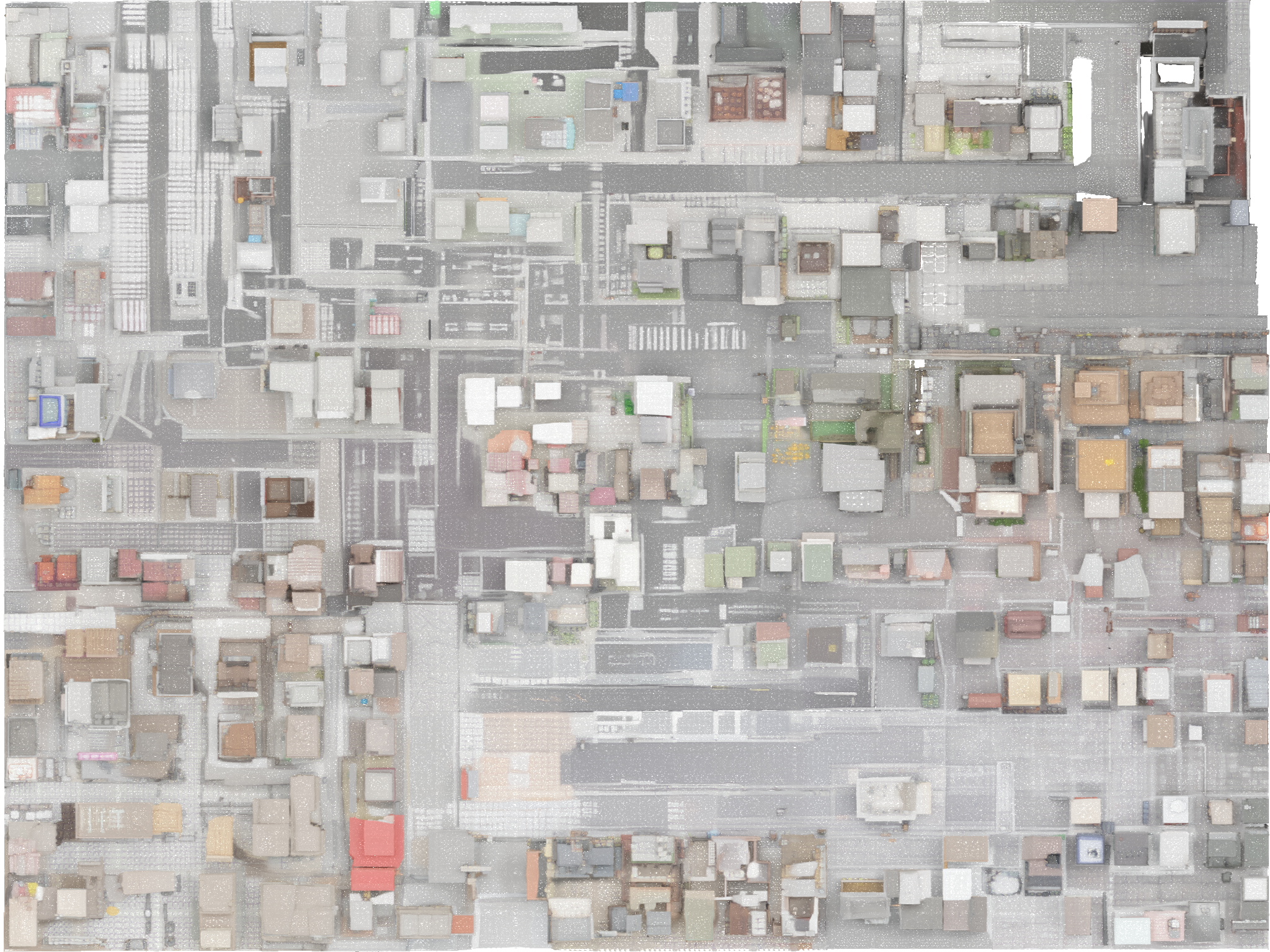} &
    \includegraphics[width=\imgw]{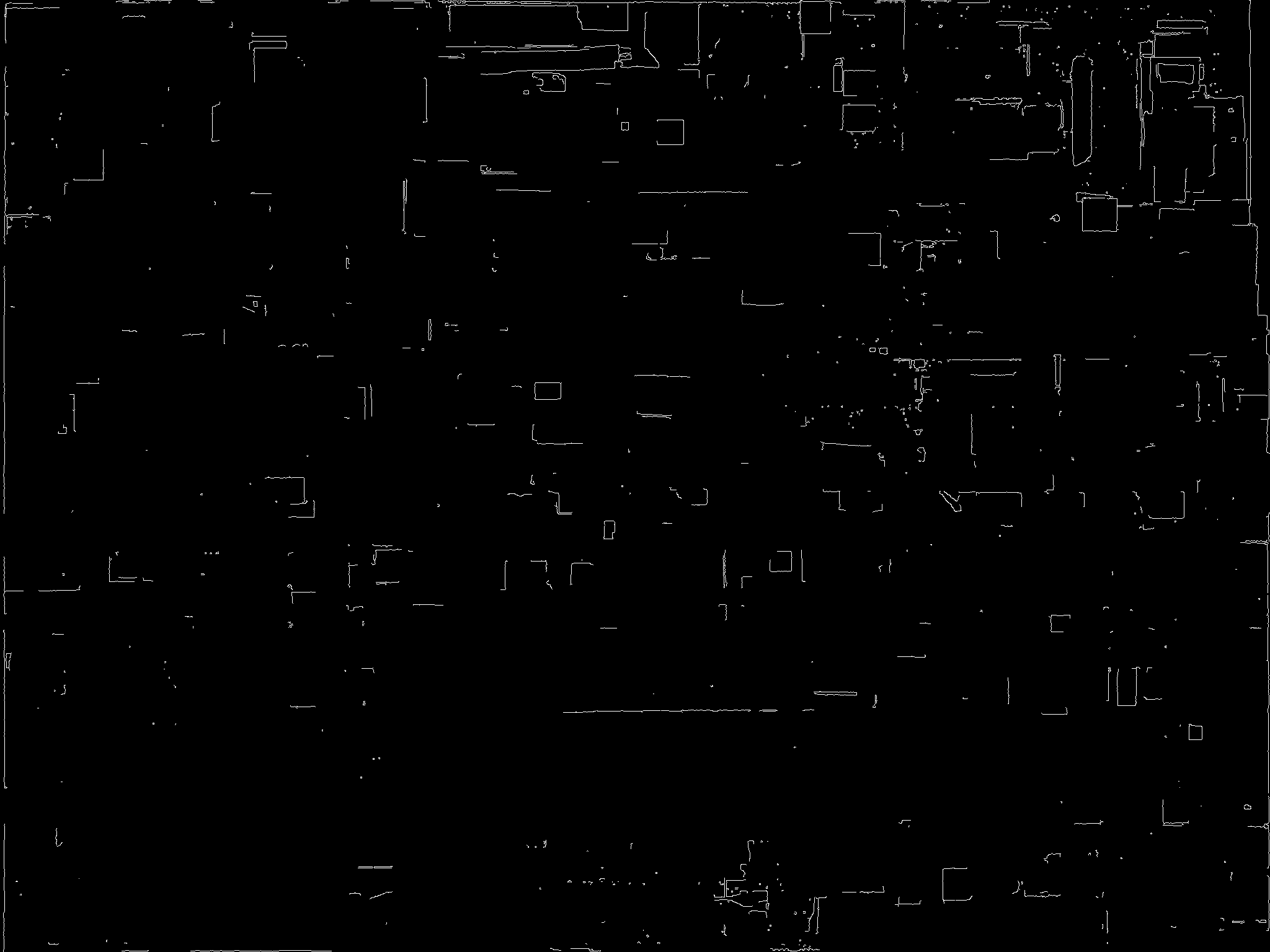} &
    \includegraphics[width=\imgw]{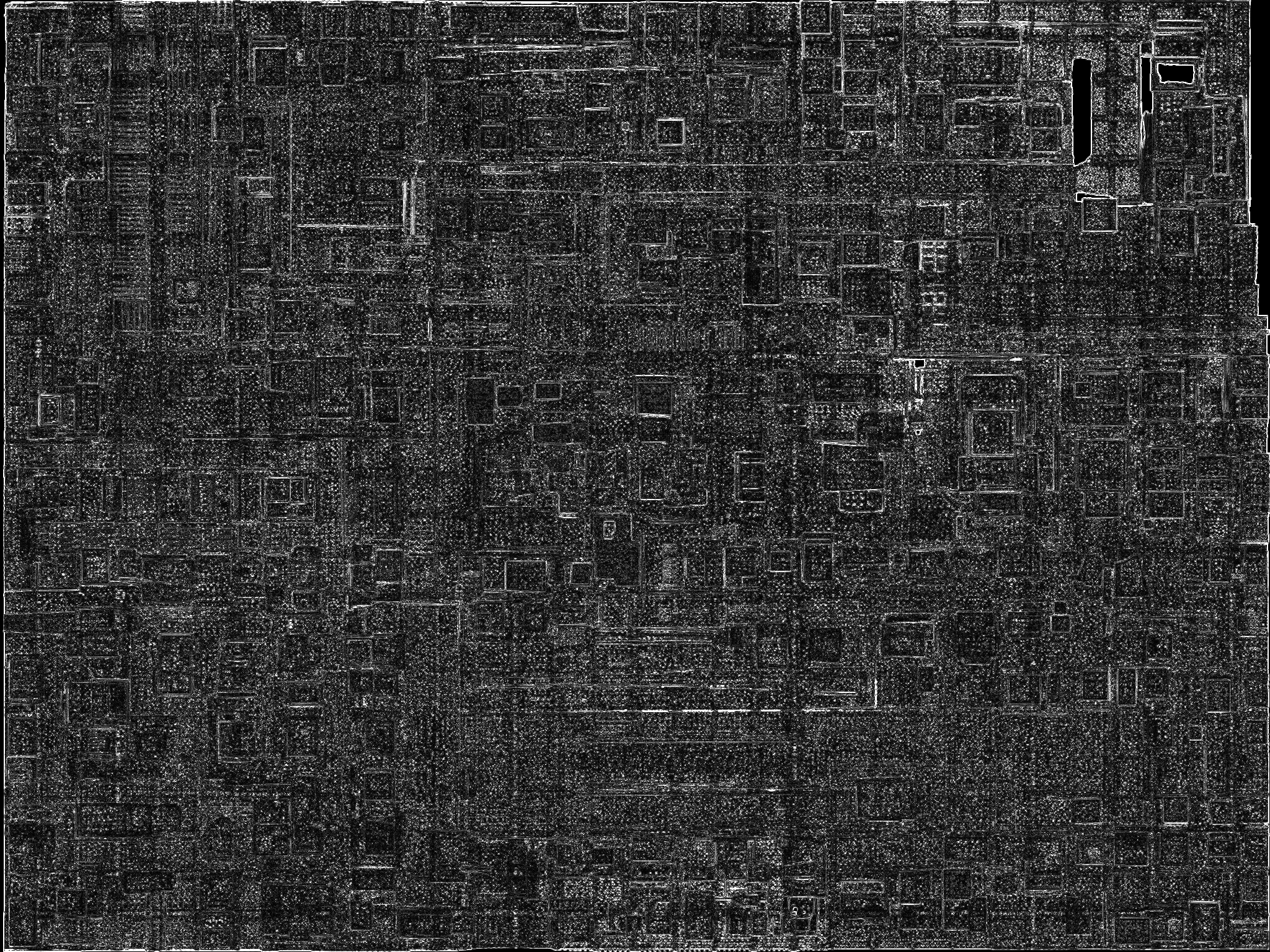} &
    \includegraphics[width=\imgw]{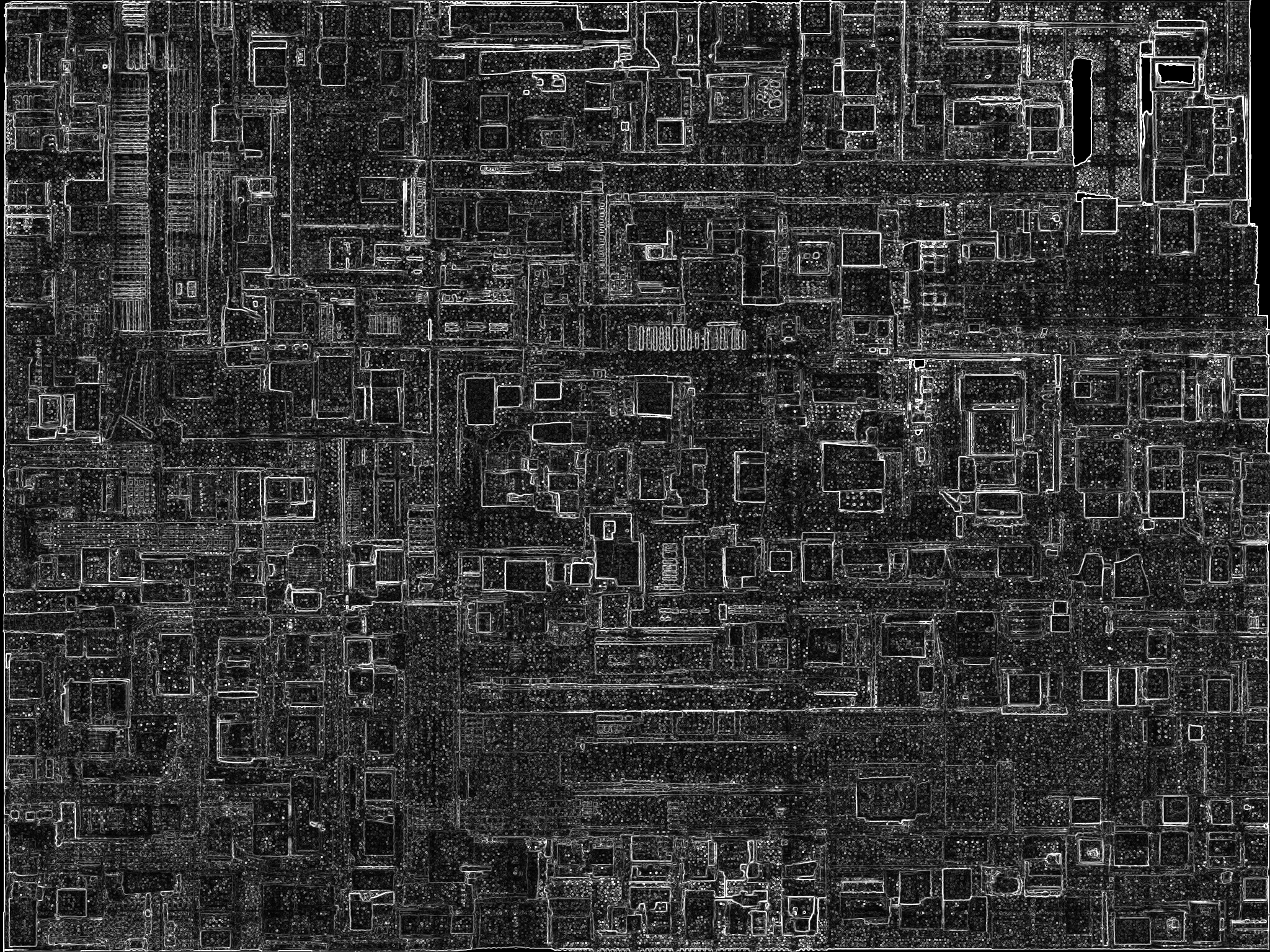}
    \\[4pt]
    \rotatebox{90}{s=16} &
    \includegraphics[width=\imgw]{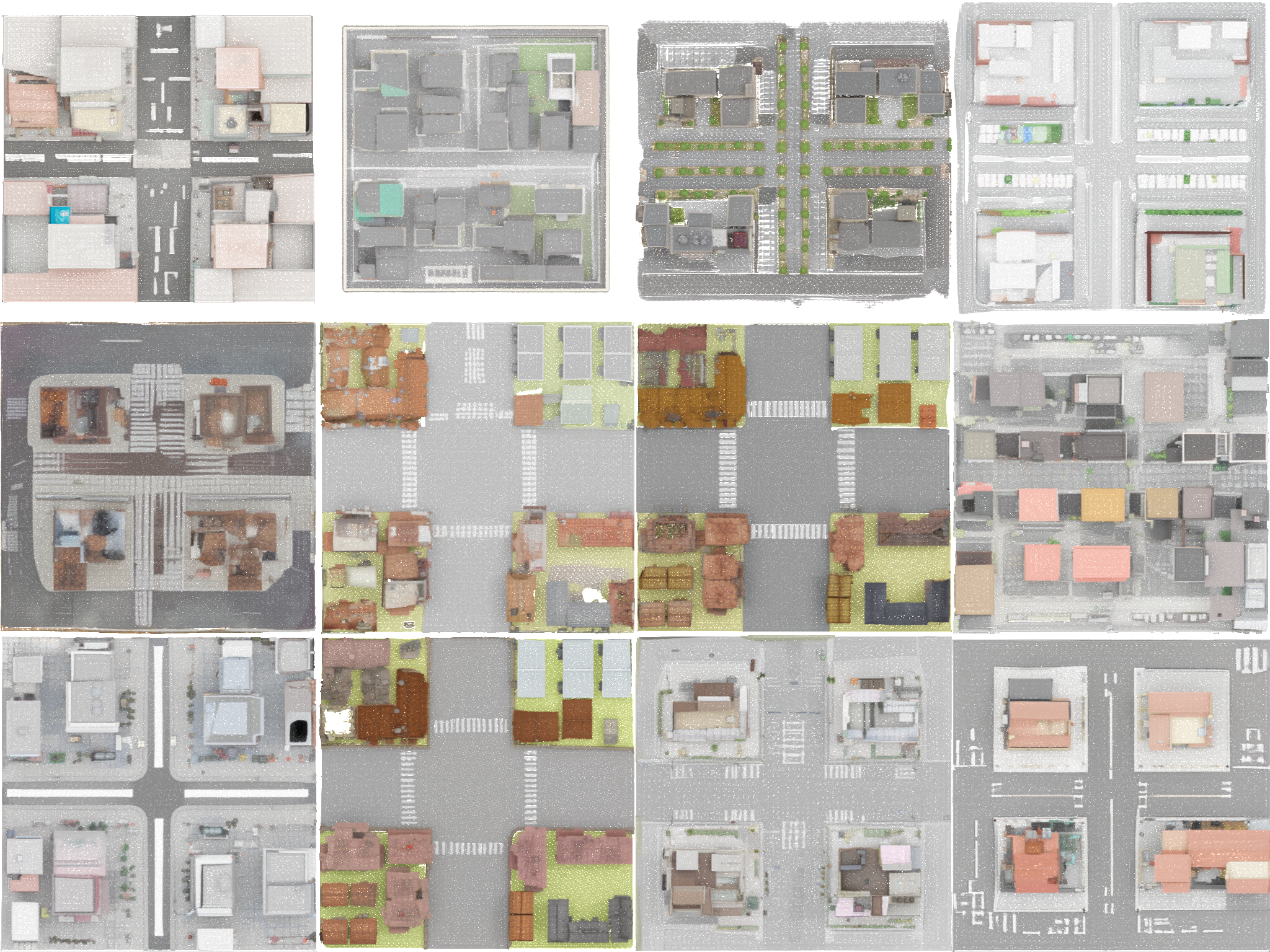} &
    \includegraphics[width=\imgw]{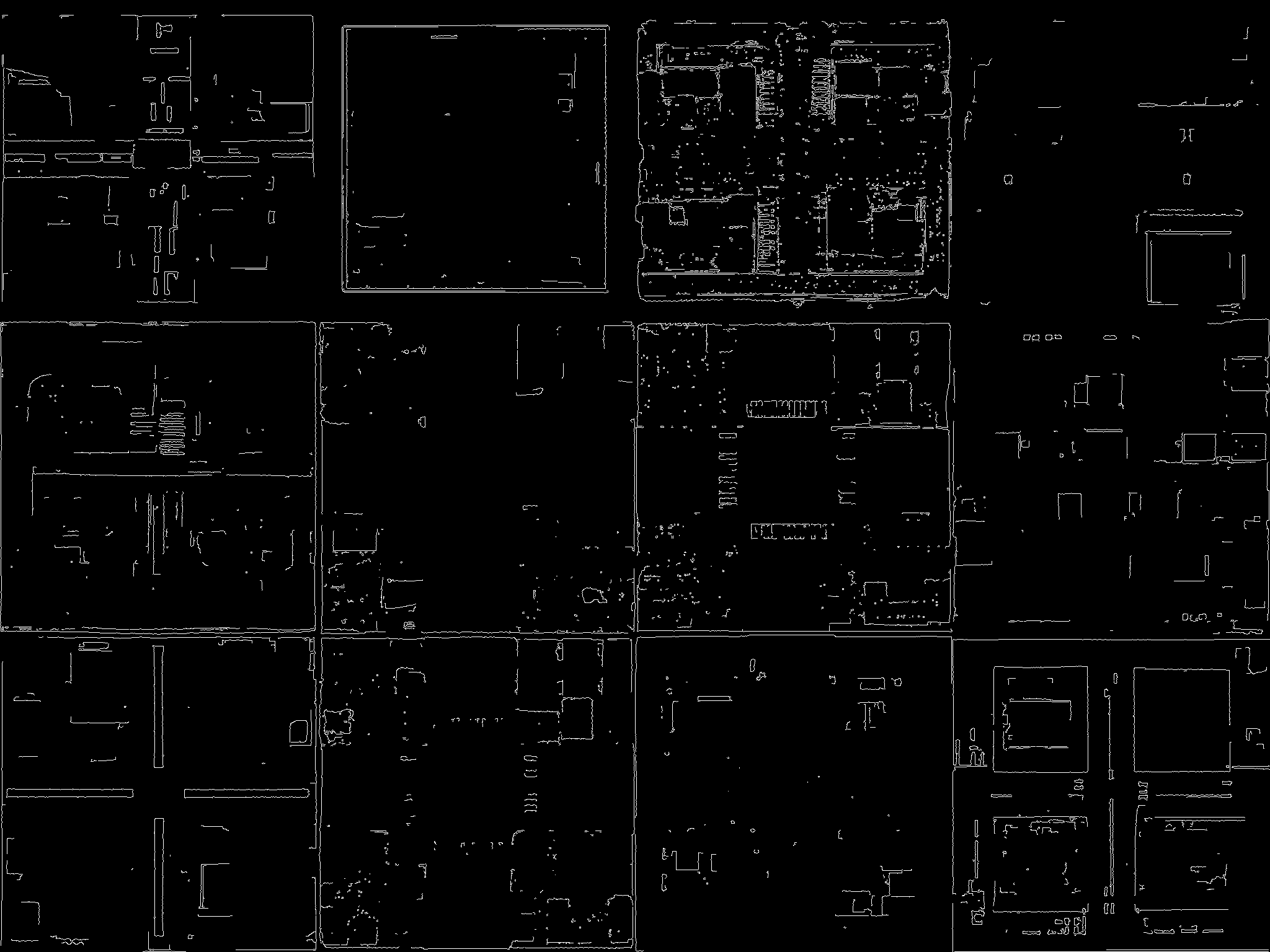} &
    \includegraphics[width=\imgw]{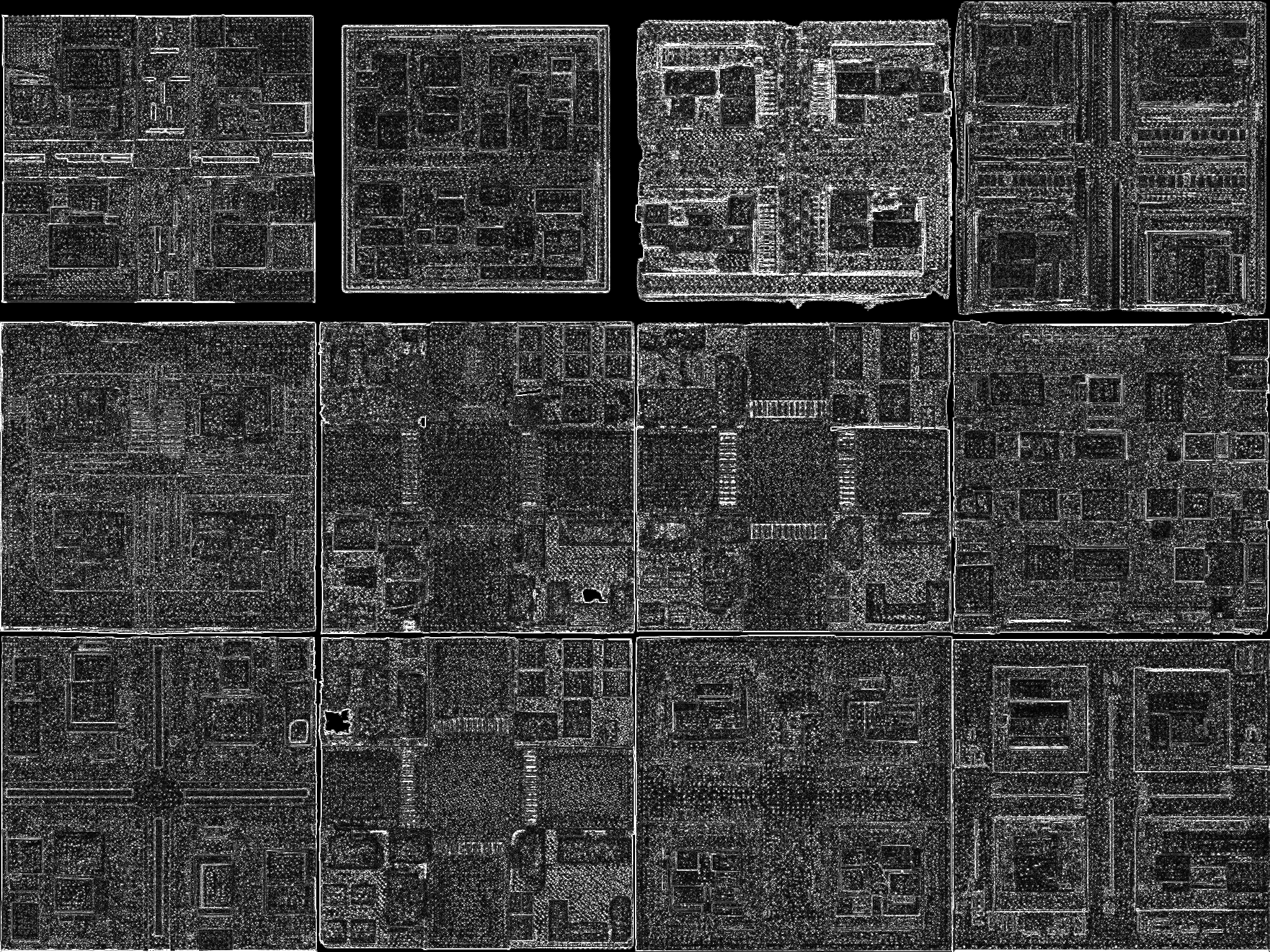} &
    \includegraphics[width=\imgw]{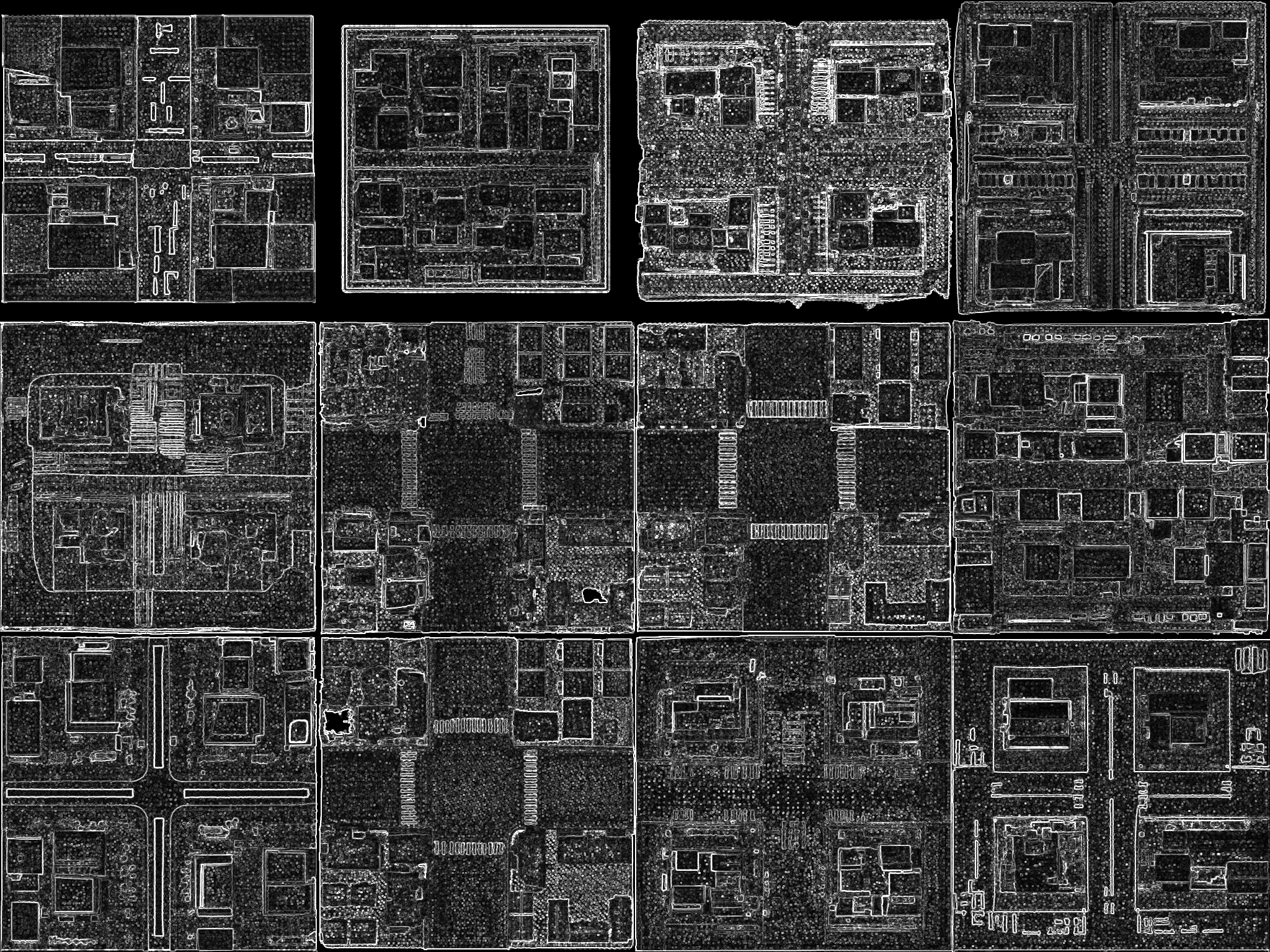}
    \\[6pt]
    &
    RGB & Canny & Laplacian & Sobel
    \\
    \end{tabular}
    \caption{Visualization of different edge detection methods. Canny best aligns with human perception of seams, while other methods highlight internal content, making them unsuitable for quantitative seam measurement.}
    \label{fig:appendix_metrics_cannyavg}
\end{figure}

We used CannyAvg to evaluate the amount of seams in our generated results. The CannyAvg value for each scene is calculated as the average pixel intensity of an orthographic top-down rendering after Canny edge detection with a $[200, 400]$ threshold. For each 4x3x1 generated result, we rendered an image at 1536x2048 resolution. To show the effectiveness of CannyAvg, we have included Canny edge detection results in \autoref{fig:appendix_metrics_cannyavg}, along with other possible edge detection methods that are less effective at highlighting seams.

\subsection{Extension to Other Base Generator}
\begin{figure}[t]
    \centering
    \newcommand{\lblw}{0.005\textwidth}   
    \newcommand{\imgw}{0.95\textwidth}   
    \begin{tabular}{%
    >{\centering\arraybackslash}m{\lblw}%
    >{\centering\arraybackslash}m{\imgw}%
    }
    \rotatebox{90}{\scriptsize(a) Terrain Generator} &
    \includegraphics[width=\imgw]{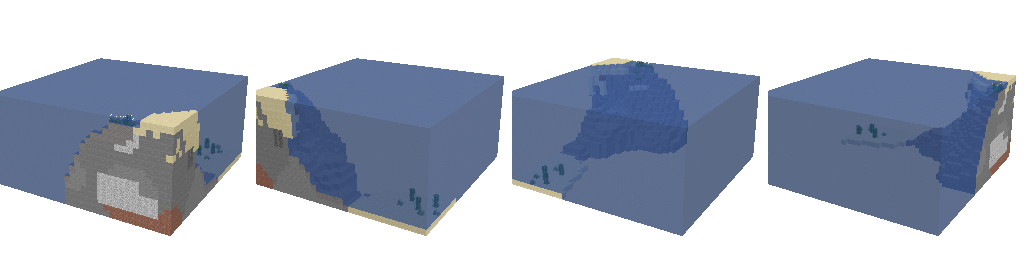}
    \\[4pt]
    \rotatebox{90}{\scriptsize(b) Structure Generator} &
    \includegraphics[width=\imgw]{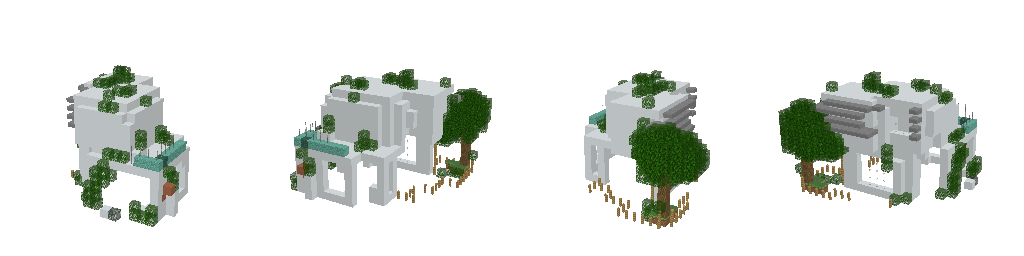}
    \\[6pt]
    &
    \\
    \end{tabular}
    \caption{Generation results of tiled diffusion applied to two different Minecraft object generators. (a) The Minecraft terrain generator is trained on 16x16x16 blocks but sampled at 32x32x16 blocks. (b) The Minecraft structure generator is trained on 16x16x16 blocks but sampled at 22x21x14 blocks.}
    \label{fig:appendix_generator}
\end{figure}

To demonstrate the extensiveness of our method, we have tested our method using two other base generators. The first is an unconditional diffusion transformer trained on Minecraft terrains and the second is an image-conditioned triplane diffusion transformer trained on crops of Minecraft structures. While both architectures only support generation of 16x16x16 blocks layout, we adapt them using our method to generate structures with sizes beyond what they were trained on. The qualitative results in \autoref{fig:appendix_generator} shows that our method keeps the generation cohesive without any visible seams.

\subsection{Extension to Image-Conditioned Generator}
\begin{figure}[t]
    \centering
    \newcommand{\lblw}{0.005\textwidth}   
    \newcommand{\imgw}{0.21\textwidth}   
    \newcommand{\resultw}{0.42\textwidth}   
    \begin{tabular}{%
    >{\centering\arraybackslash}m{\lblw}%
    >{\centering\arraybackslash}m{\imgw}%
    >{\centering\arraybackslash}m{\imgw}%
    >{\centering\arraybackslash}m{\resultw}%
    }
    \rotatebox{90}{City} &
    \includegraphics[width=\imgw]{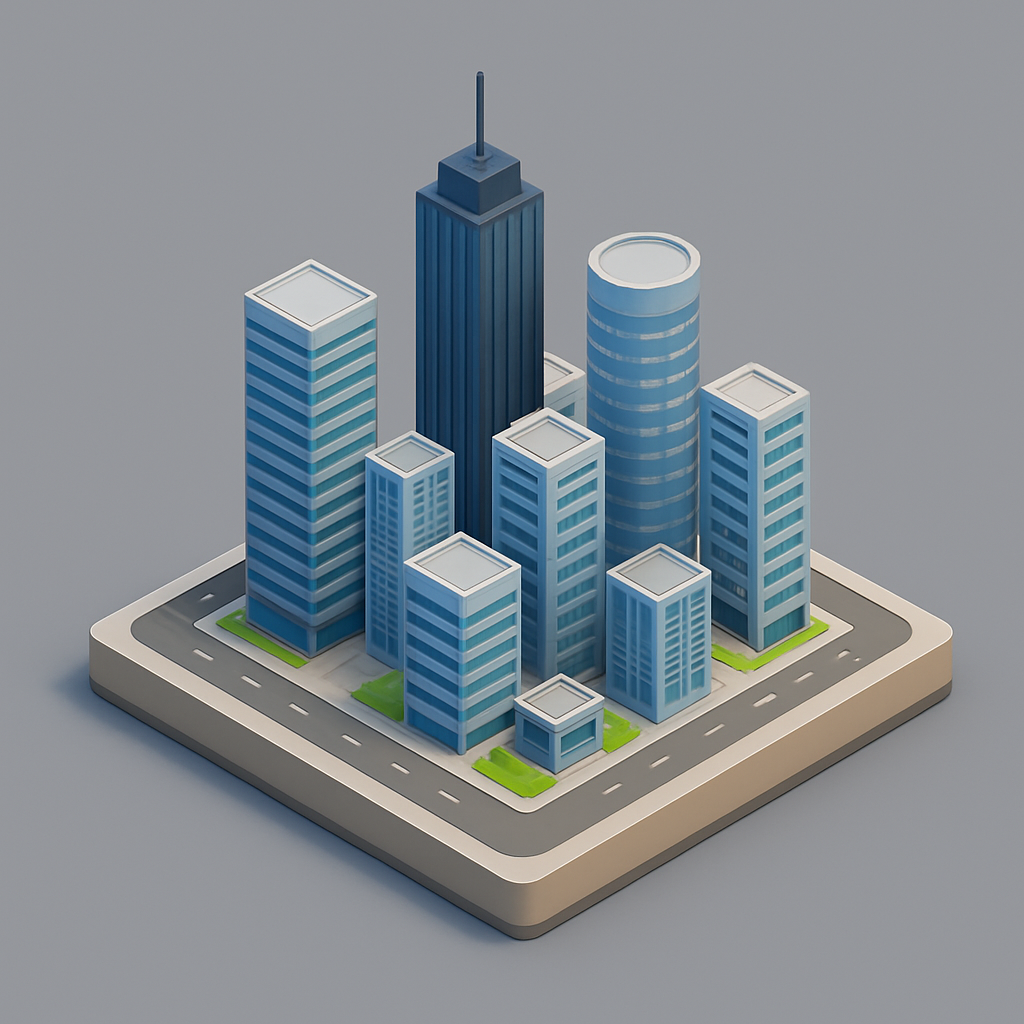} &
    \includegraphics[width=\imgw]{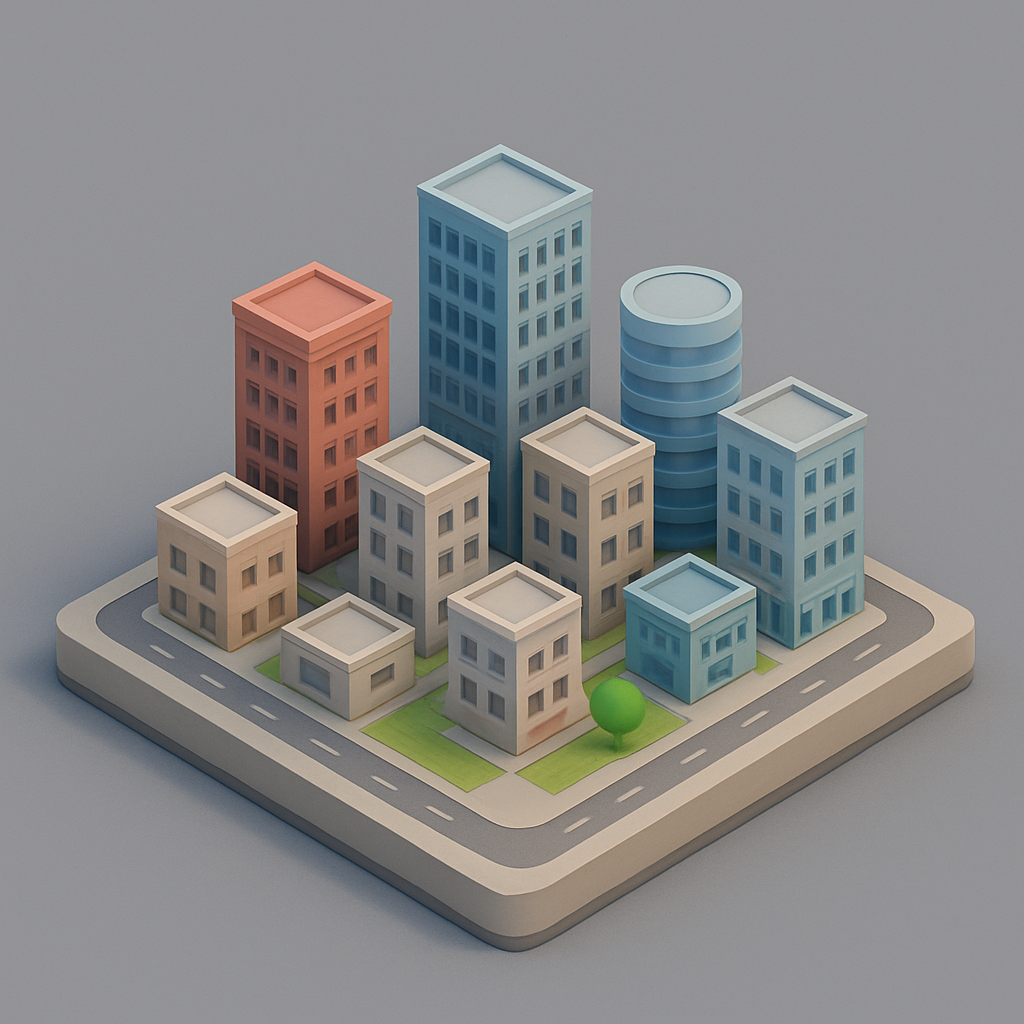} &
    \includegraphics[width=\resultw]{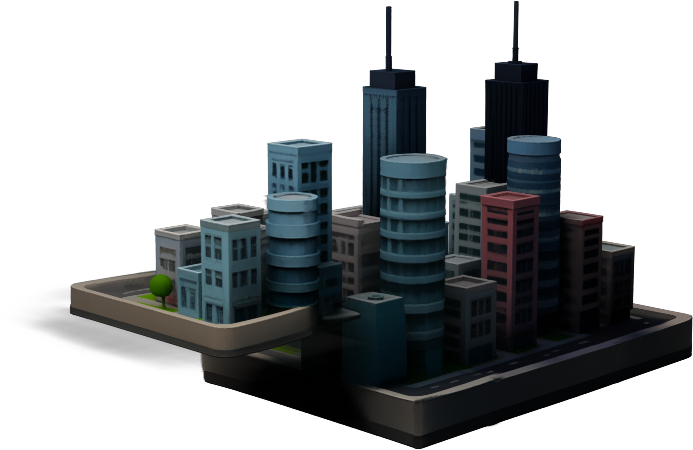}
    \\[4pt]
    \rotatebox{90}{Minecraft} &
    \includegraphics[width=\imgw]{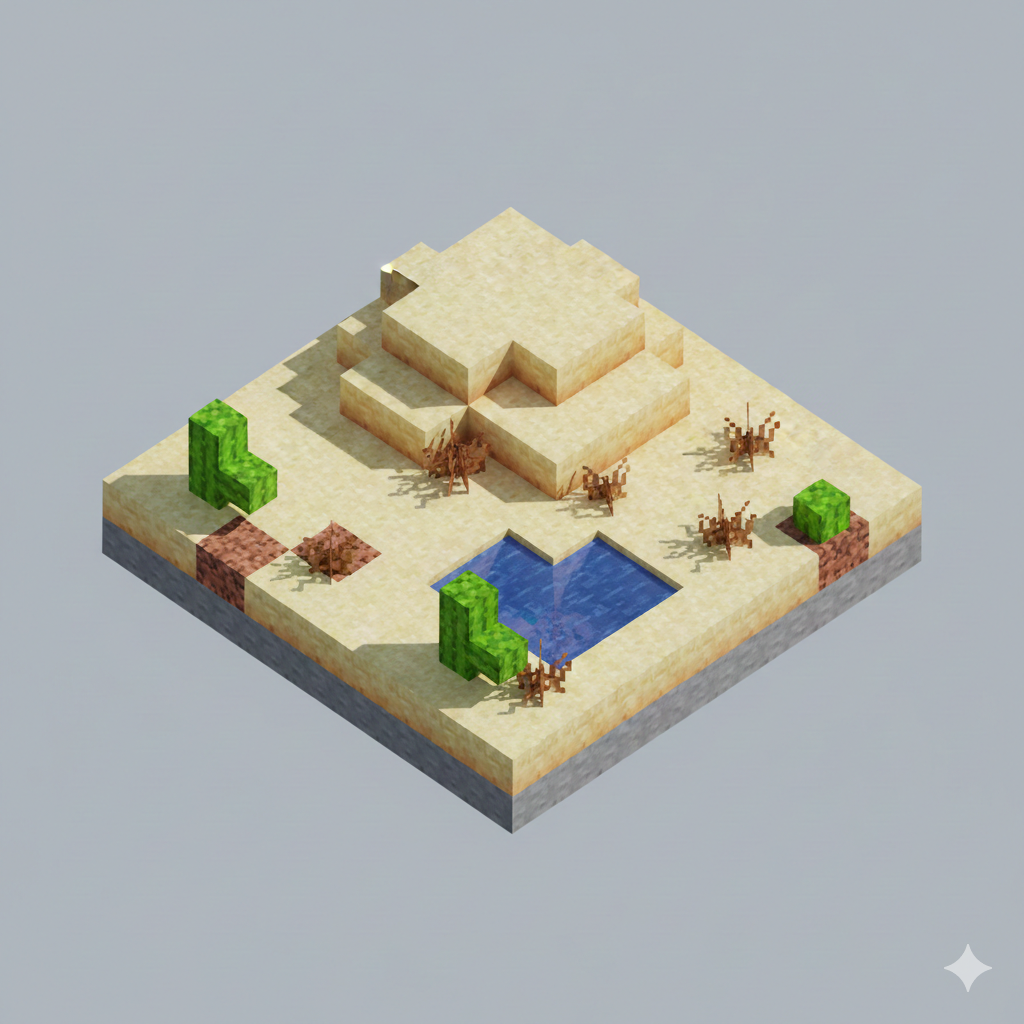} &
    \includegraphics[width=\imgw]{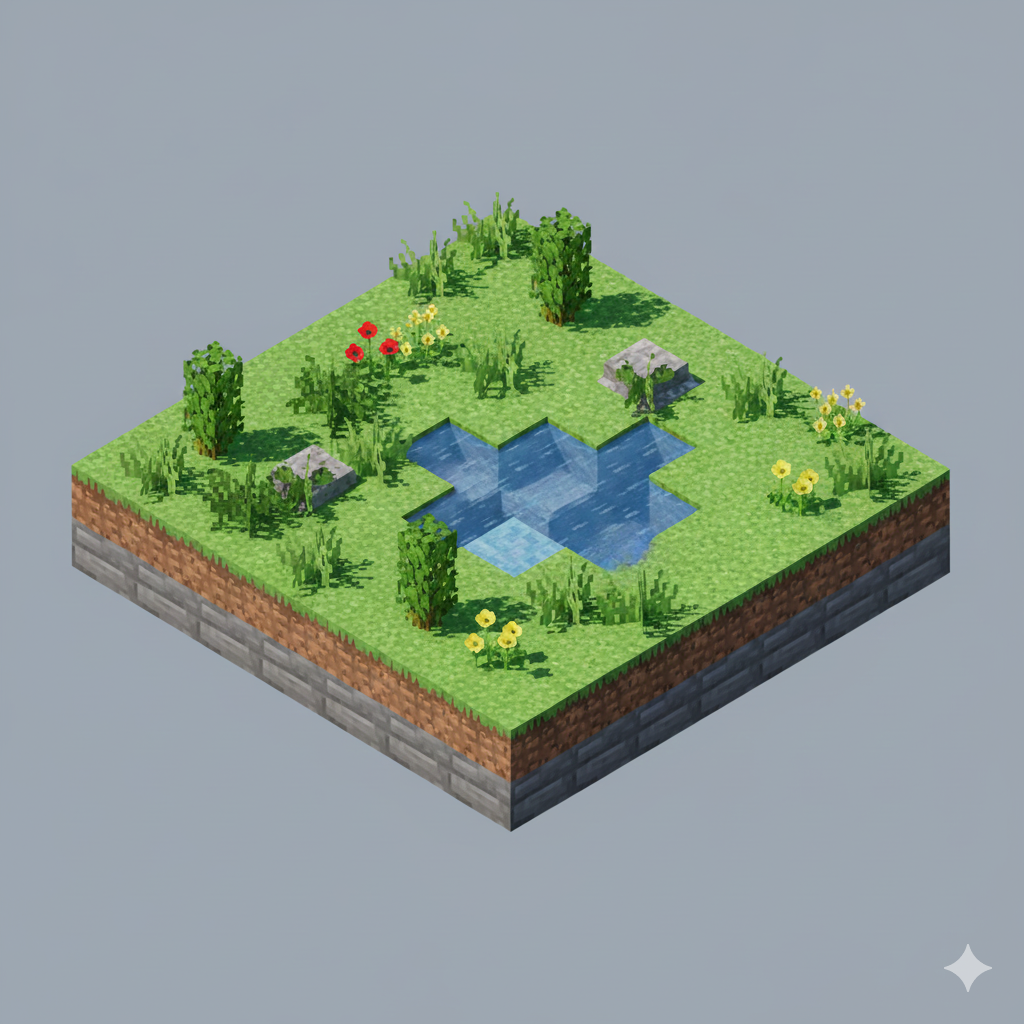} &
    \includegraphics[width=\resultw]{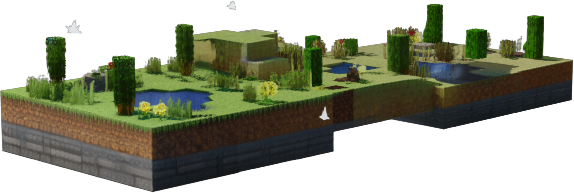}
    \\[6pt]
    \rotatebox{90}{Village} &
    \includegraphics[width=\imgw]{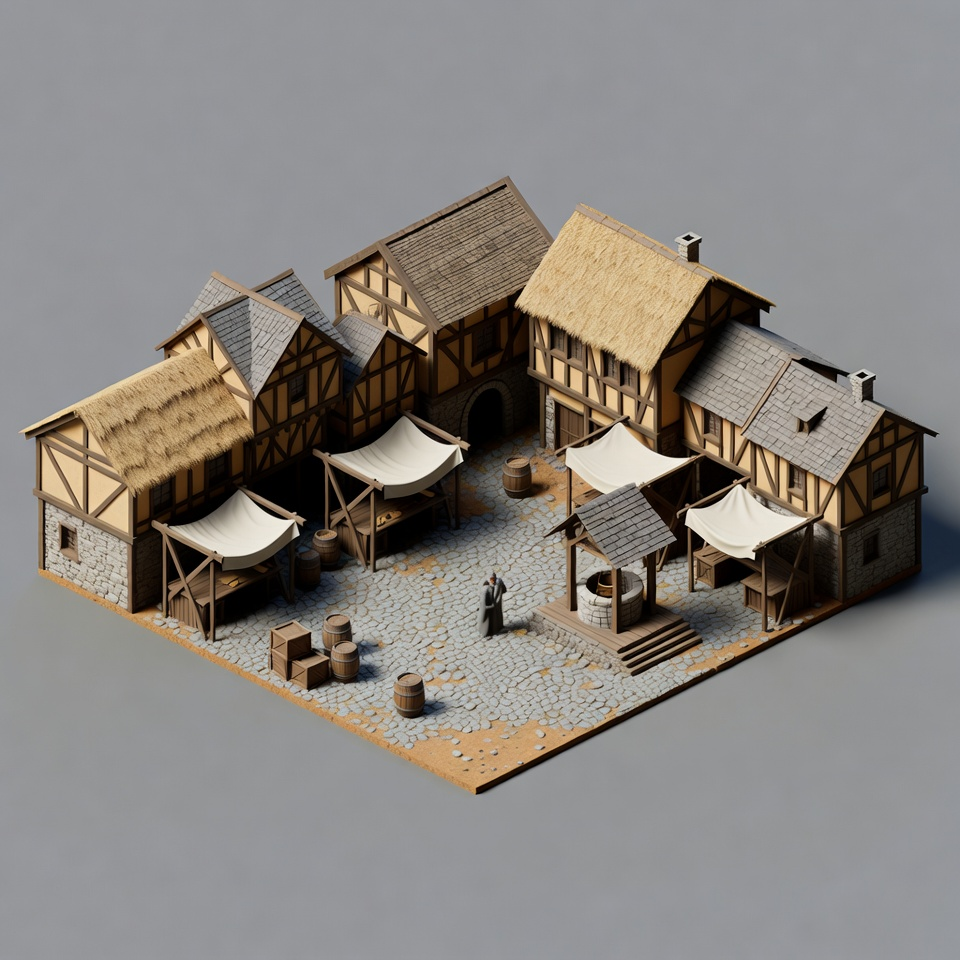} &
    \includegraphics[width=\imgw]{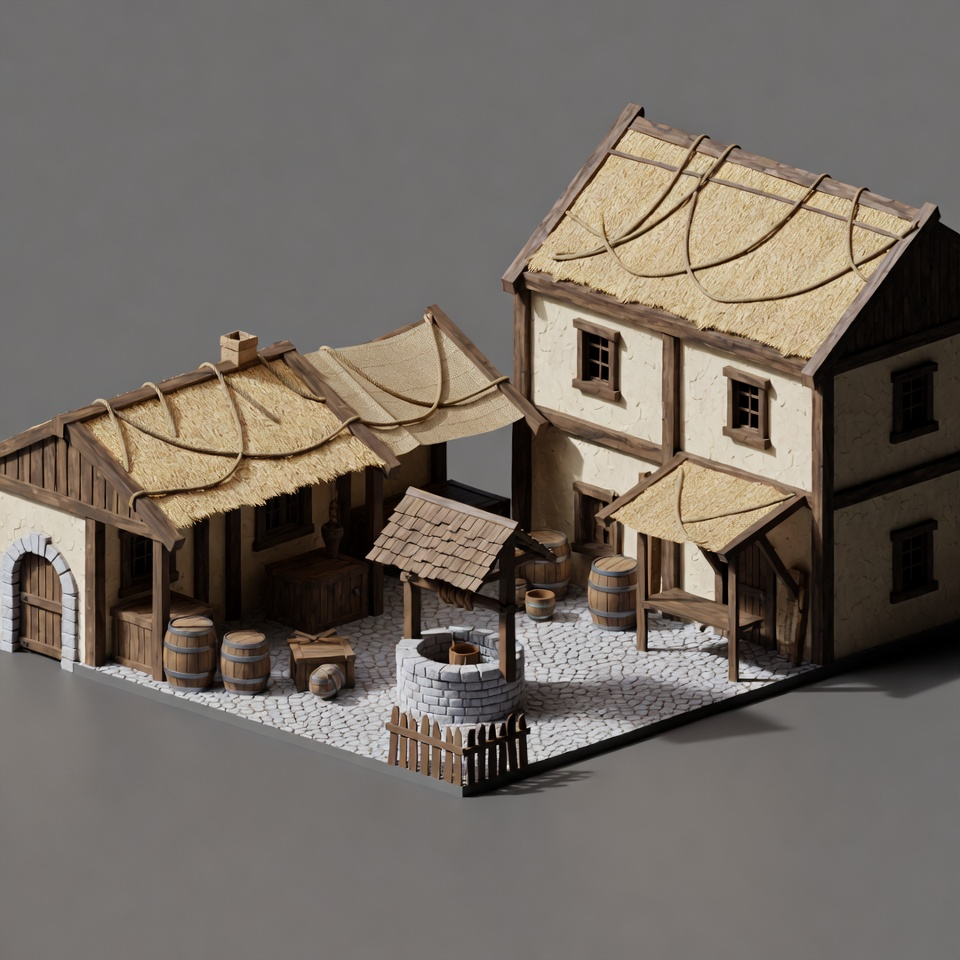} &
    \includegraphics[width=\resultw]{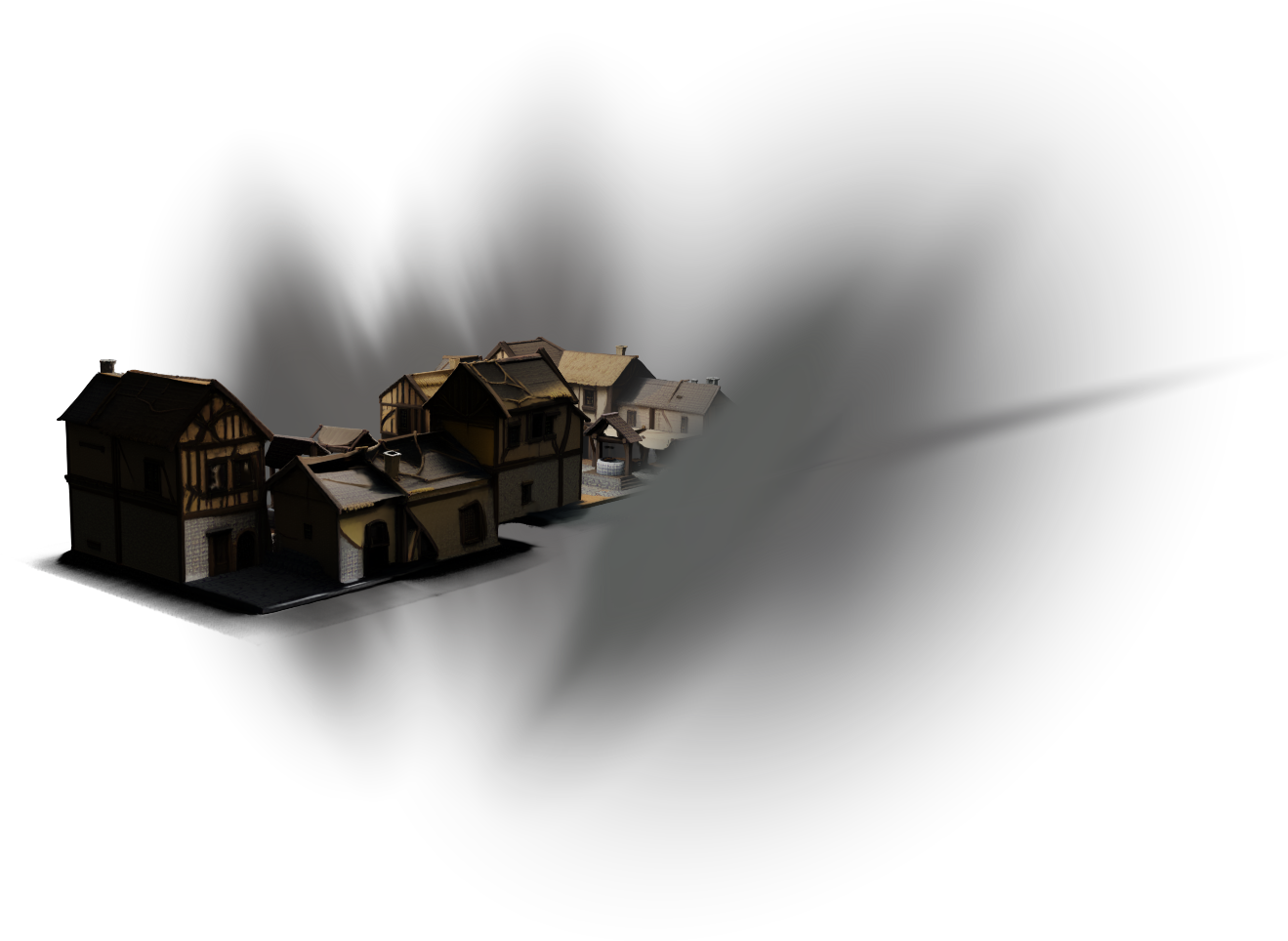}
    \\[6pt]
    &
    Image Left & Image Right & Generated Result
    \\
    \end{tabular}
    \caption{Visualization of applying TRELLISWorld to image-conditioned generator. We generated 2x1x1 chunk which was split to 3 tiles (left, right, and middle tile) from 2 conditional images. The middle image was chosen arbitrarily from left or right images. In all generations give inconsistent floor levels. The "Village" example shows significant artifact.}
    \label{fig:appendix_image_condition}
\end{figure}

We argue that our method is best applied to a text-conditioned object generator. Applying our method to image-conditioned object generators often result in unaligned floor level, which is an indicator of unsuccessful blending. We show generation results using the image-conditioned TRELLIS model in \autoref{fig:appendix_image_condition}.

We suspect that this discrepancy between text-conditioned and image-conditioned model lies in the difference in their underlying learned conditional distributions of the score models $\theta_{img}, \theta_{text}$. When the model $\theta_{img}, \theta_{text}$ is trained on the same set of 3D object but with different conditional label $c_{img}, c_{text}$. They have the same marginalized distribution $q_{\theta_{img}}(x_0) = \int q_{\theta_{img}}(x_0\mid c_{img})p(c_{img})dc_{img} = \int q_{\theta_{text}}(x_0\mid c_{text})p(c_{text})dc_{text} = q_{\theta_{text}}(x_0)$. However, their conditional distribution is different: $q_{\theta_{text}}(x_0\mid c_{text}=\text{specific-text}) \neq q_{\theta_{img}}(x_0\mid c_{img}=\text{specific-img})$ even when the given "specific-text" and "specific-img" are associated with the same 3D object. Specifically, $q_{\theta_{text}}(x_0\mid c_{text})$ might be more diffuse compared to $q_{\theta_{img}}(x_0\mid c_{img})$, which makes the text-conditioned model more robust to out-of-data-distribution queries of the score function.

\subsection{Additional Qualitative Results}

\begin{figure}[!t]
  \centering
  \includegraphics[trim=0 0 0 0,clip,width=0.99\linewidth]{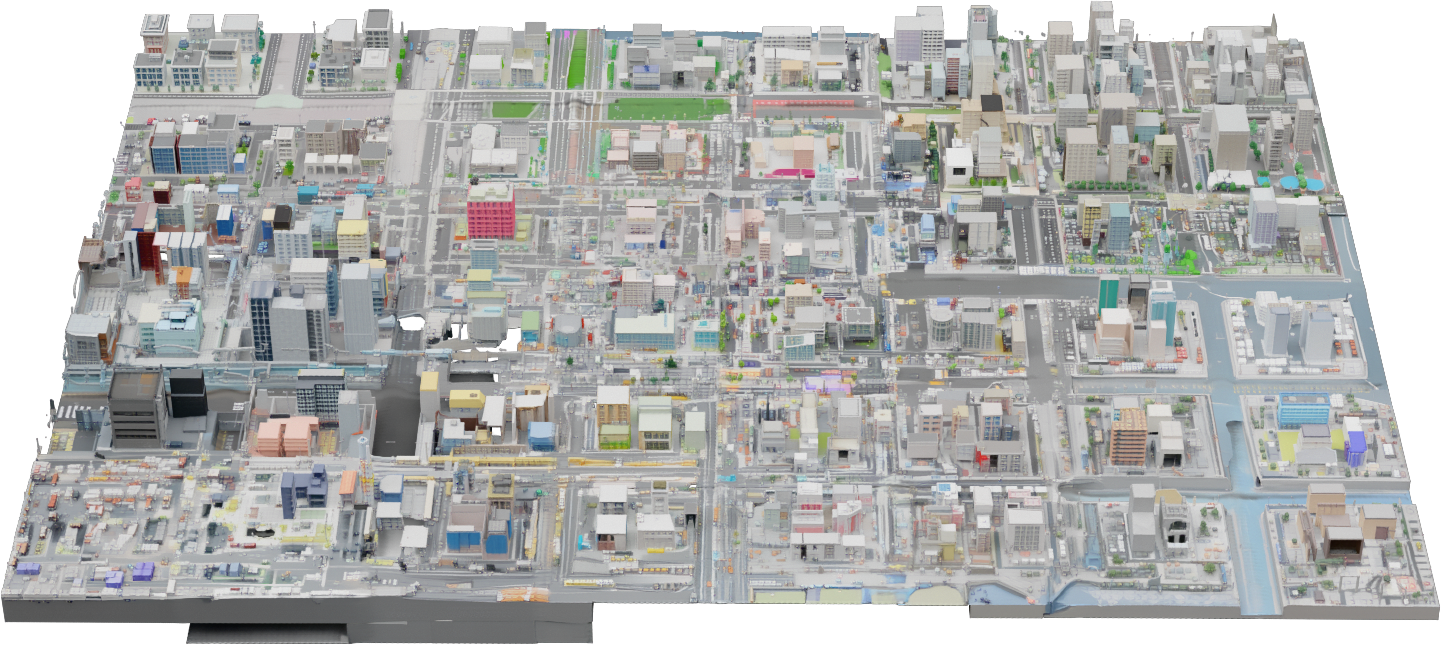}
  \includegraphics[trim=0 0 0 0,clip,width=0.99\linewidth]{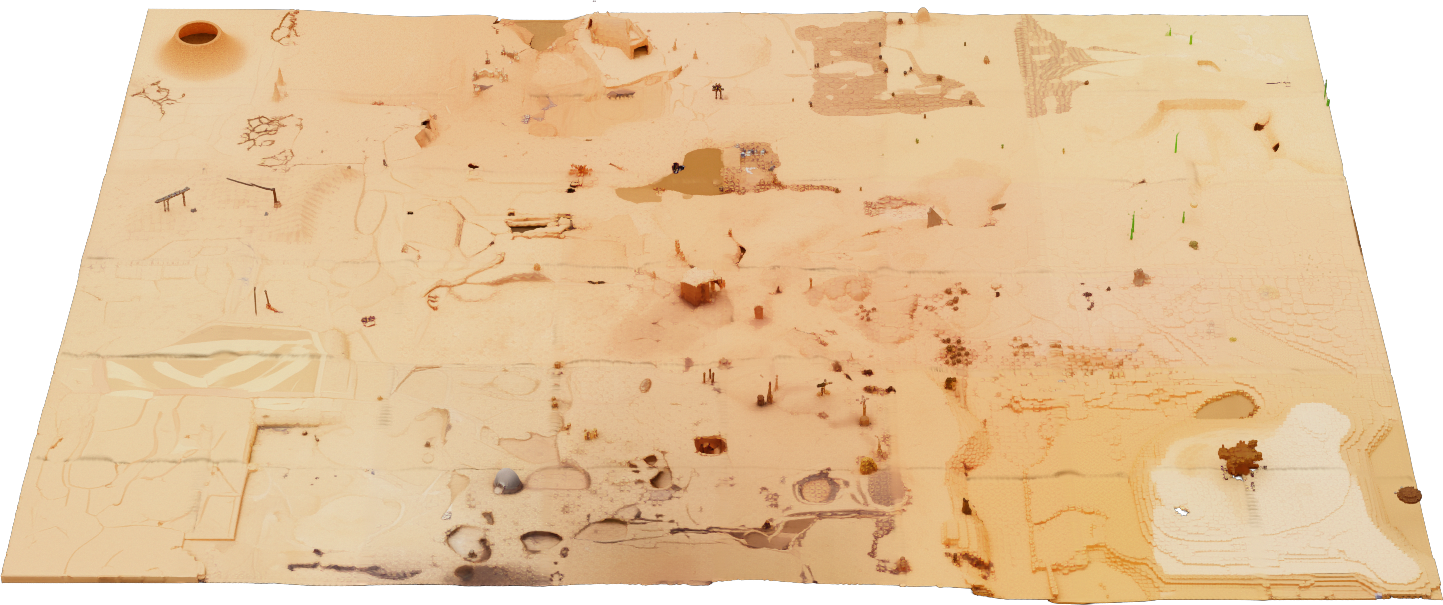}
  \includegraphics[trim=0 0 0 0,clip,width=0.99\linewidth]{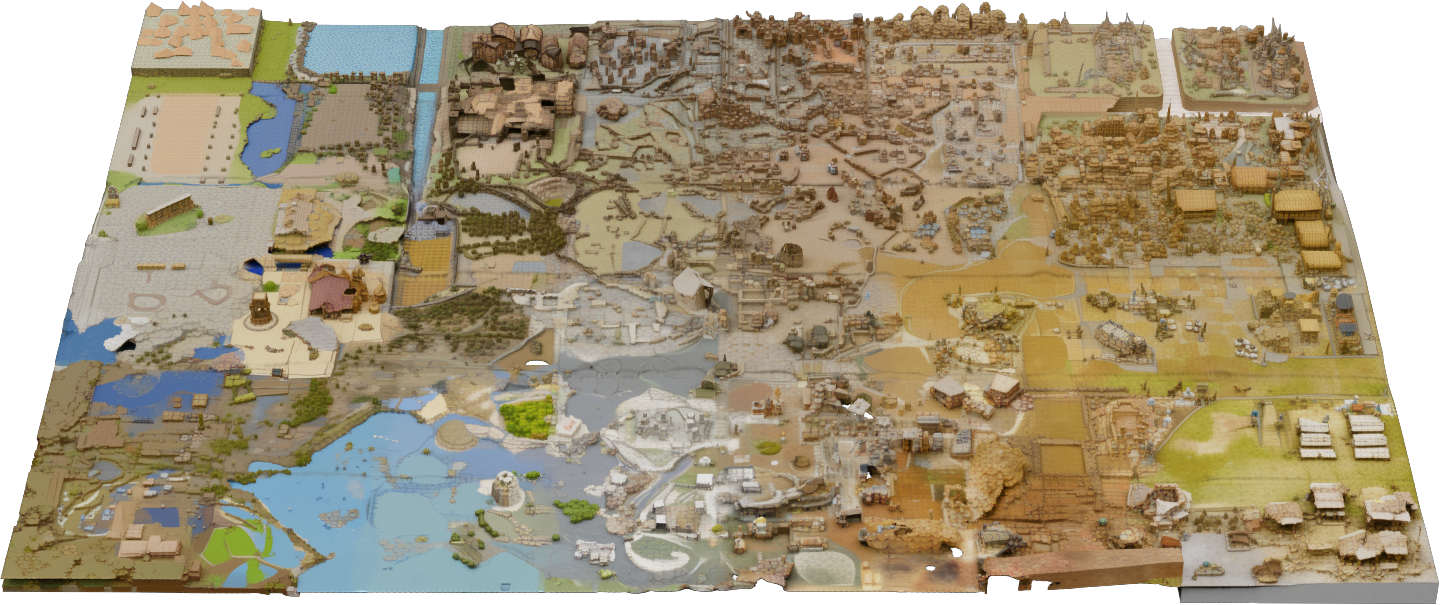}
  \caption{Additional (not cherry-picked) 4x3x1 chunk Tiled Gaussian Splatting Generation Results: city, desert, medieval.}
  \label{fig:appendix_additional_1}
\end{figure}

\begin{figure}[!t]
  \centering
  \includegraphics[trim=0 0 0 0,clip,width=0.99\linewidth]{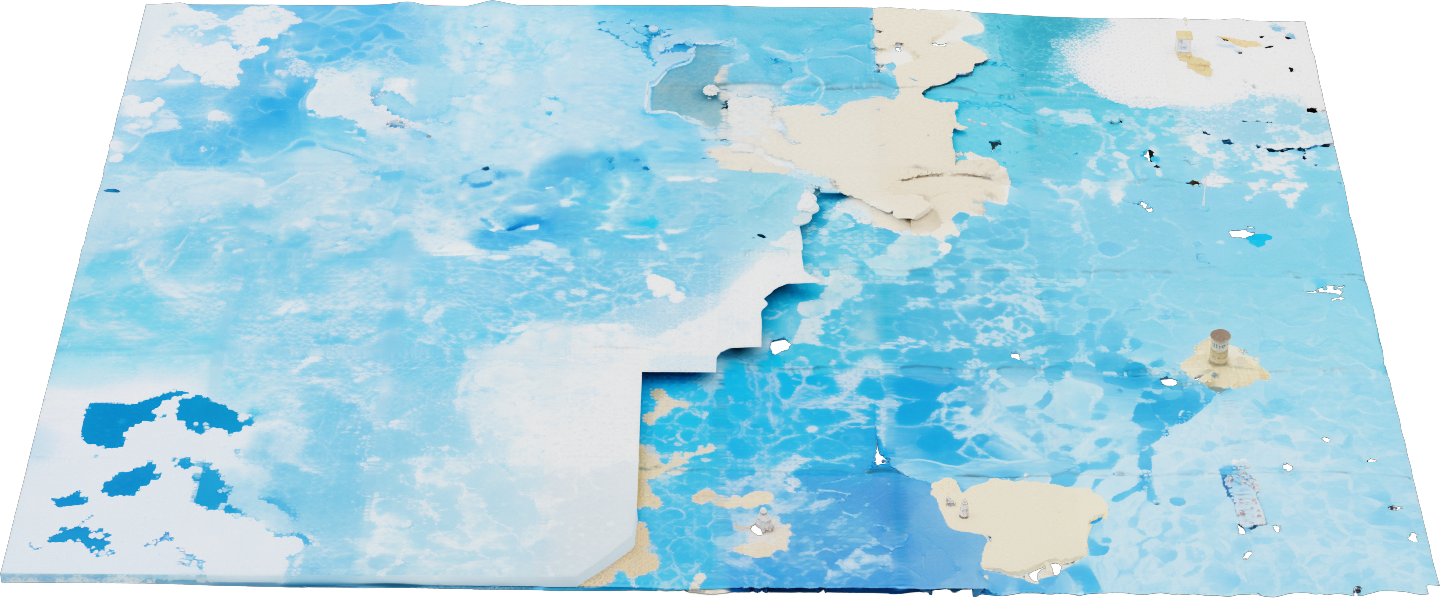}
  \includegraphics[trim=0 0 0 0,clip,width=0.99\linewidth]{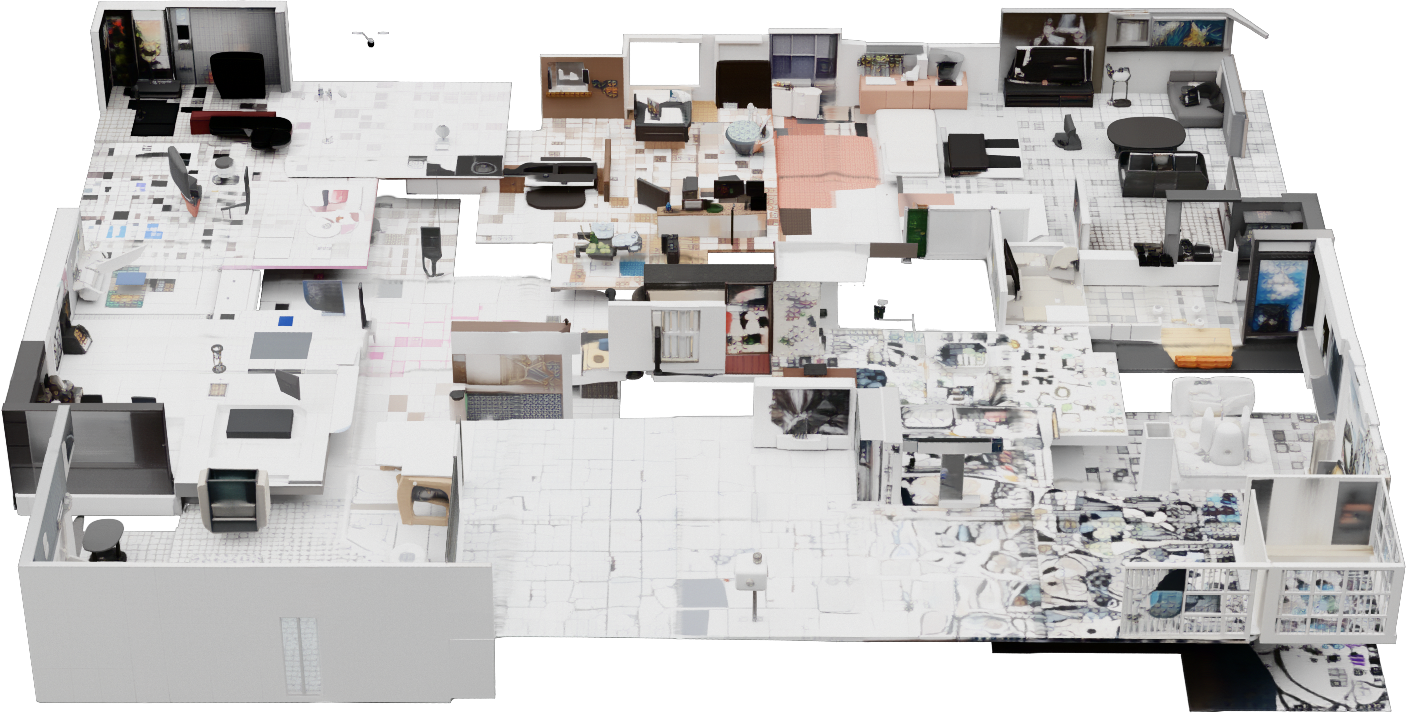}
  \includegraphics[trim=0 0 0 0,clip,width=0.99\linewidth]{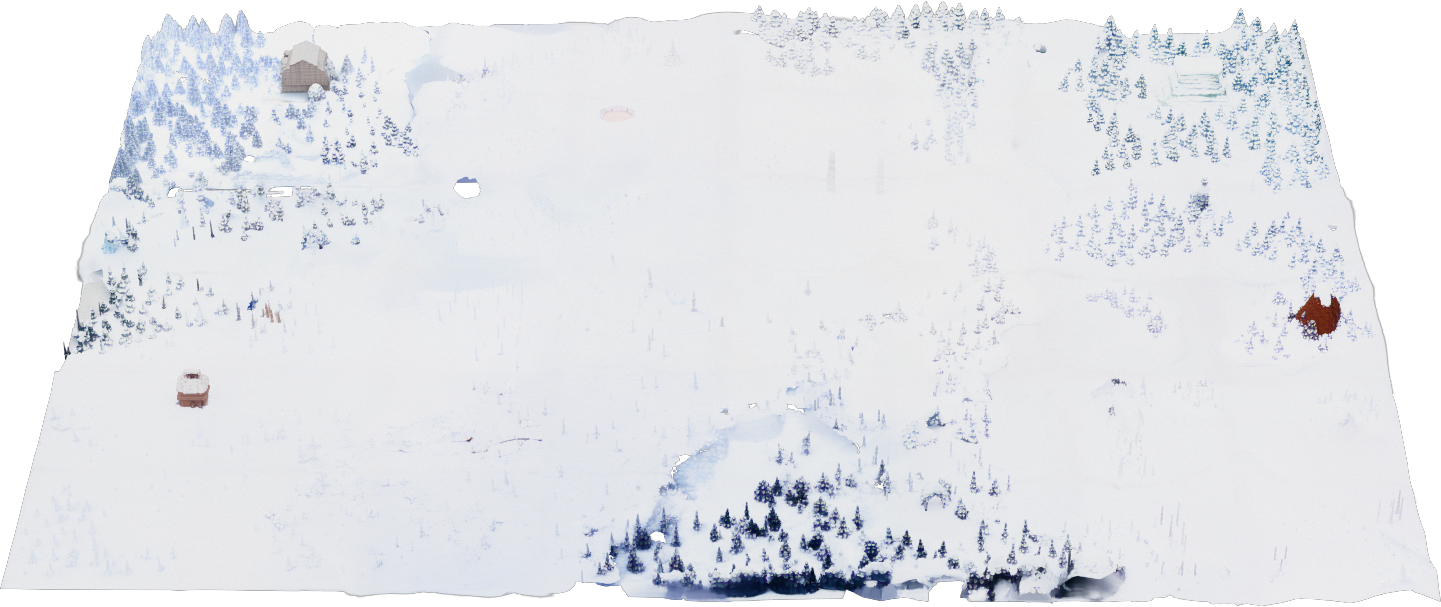}
  \caption{Additional (not cherry-picked) 4x3x1 chunk Tiled Gaussian Splatting Generation Results: ocean, room, winter.}
  \label{fig:appendix_additional_2}
\end{figure}

\autoref{fig:appendix_additional_1} and \autoref{fig:appendix_additional_2} shows additional generation result of TRELLISWorld.

\subsection{Disclosure}
We made use of LLMs to polish writing. We made sure that our input text to LLMs will not be used for training purposes.

\end{document}